\pdfoutput=1

\documentclass[11pt]{article}

\usepackage[]{acl}
\usepackage[para]{footmisc}

\usepackage{times}
\usepackage{latexsym}
\usepackage{algorithmic}
\usepackage[algoruled]{algorithm2e}

\usepackage[T1]{fontenc}

\usepackage[utf8]{inputenc}

\usepackage{microtype}

\usepackage{inconsolata}

\usepackage{graphicx}

\usepackage{amsmath}
\usepackage{amssymb}
\usepackage{booktabs}
\usepackage{authblk} 

\usepackage{color, booktabs,subcaption,amsfonts,dcolumn}
\usepackage{xcolor}
\usepackage{mdframed}
\newmdenv[leftmargin=5pt,rightmargin=5pt,backgroundcolor=gray!20,innertopmargin=5pt,innerbottommargin=5pt]{coloredquote}

\newenvironment{smallermdframed}{
  \begin{coloredquote} \scriptsize 
}{\end{coloredquote}}

\usepackage{amsfonts}
\usepackage{graphicx}

\usepackage{multirow,diagbox,makecell}
\usepackage{tikz}

\usepackage{pifont}
\usepackage{booktabs}

\usepackage{capt-of}
\usepackage{caption}
\usepackage{xcolor}
\usepackage{cleveref}

\newcommand{\oursB}{\texttt{GUI-Bee}\xspace}

\def\logoF{\makebox[0pt][l]{\hspace{-3pt}\raisebox{-0.3ex}{\includegraphics[height=18pt]{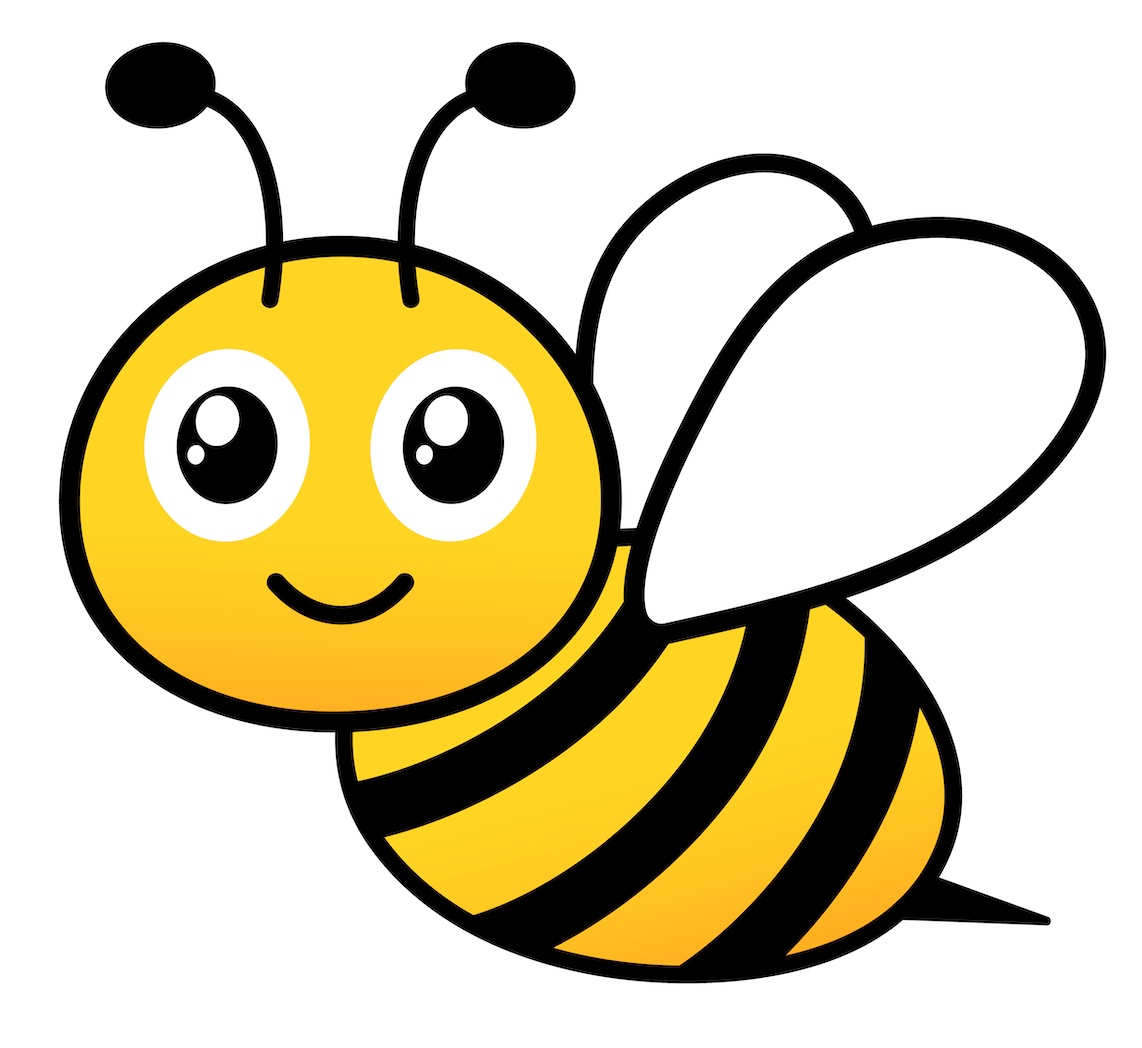}}}}
\title{
\oursB~\logoF~~~~~: Align GUI Action Grounding to Novel Environments via Autonomous Exploration
}

\author{Yue Fan \textsuperscript{1}, Handong Zhao \textsuperscript{2}, Ruiyi Zhang \textsuperscript{2}, Yu Shen \textsuperscript{2}, Xin Eric Wang  \textsuperscript{$*$1}, Gang Wu \textsuperscript{$*$2}\\
\textsuperscript{1} University of California, Santa Cruz
\textsuperscript{2} Adobe Research
\\
\small{
    \{yfan71, xwang366\}@ucsc.edu; \{hazhao, ruizhang, shenyu, gawu\}@adobe.com 
 }

}

\makeatletter
\AtBeginDocument{
\let\@oldmaketitle\@maketitle
\renewcommand{\@maketitle}{\@oldmaketitle
  \vspace{-27pt}
  \includegraphics[width=\linewidth]{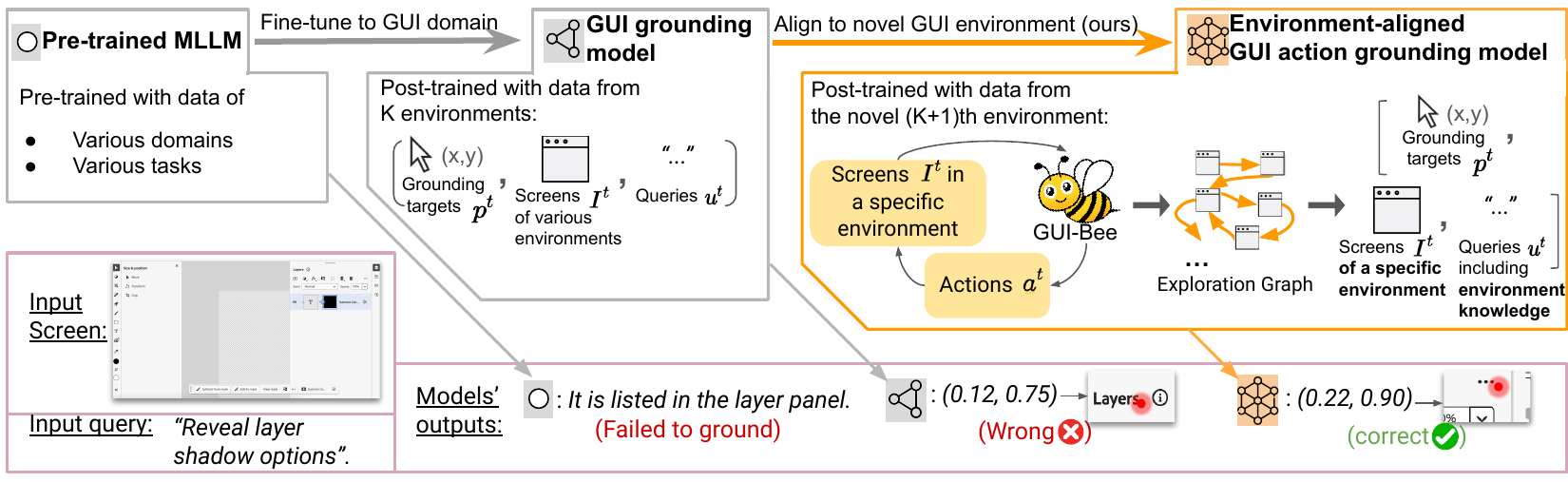}
  \vspace{-18pt}
  \captionof{figure}{
Our \textcolor{orange}{environment-aligned GUI action grounding model}, based on GUI grounding models from \textcolor{gray}{prior works}, is aligned to novel environments. Our proposed alignment process includes first exploring the specific novel environment with the GUI-Bee agent to generate the exploration graph and then fine-tuning the model with the data from the exploration graph.
In the \textcolor{purple}{inference example at the bottom}, the models encounter a query requiring knowledge of an environment-specific action outcome, which highlights the importance of the proposed alignment process. 
  }
  \label{fig:1}
  \vspace{17pt}
 }
}
\makeatother

\begin{document}
\maketitle

\renewcommand*{\thefootnote}{*}
\footnotetext[2]{
\footnotesize Co-advising. This work was partly performed when the first author interned at Adobe Research.}
\renewcommand*{\thefootnote}{\arabic{footnote}}


\begin{abstract}
Graphical User Interface (GUI) action grounding is a critical step in GUI automation that maps language instructions to actionable elements on GUI screens. Most recent works of GUI action grounding leverage large GUI datasets to fine-tune MLLMs. However, the fine-tuning data always covers limited GUI environments, and we find the performance of the resulting model deteriorates in novel environments. We argue that the GUI grounding models should be further aligned to the novel environments to reveal their full potential, when the inference is known to involve novel environments, i.e., environments not used during the previous fine-tuning. To realize this, we first propose GUI-Bee, an MLLM-based autonomous agent, to collect high-quality, environment-specific data through exploration and then continuously fine-tune GUI grounding models with the collected data. Our agent leverages a novel Q-value-Incentive In-Context Reinforcement Learning (Q-ICRL) method to optimize exploration efficiency and data quality. Additionally, we introduce NovelScreenSpot, a benchmark for testing how well the data can help align GUI action grounding models to novel environments and demonstrate the effectiveness of data collected by GUI-Bee in the experiments. Furthermore, we conduct an ablation study to validate the Q-ICRL method in enhancing the efficiency of GUI-Bee. Project page: \url{https://gui-bee.github.io}.
\end{abstract}


\section{Introduction}

GUI action grounding maps natural language instructions to specific executable elements or locations on a GUI screen. It is critical as being a key step adopted by GUI agents to assist humans in complex digital environments, where GUI agents usually generate step-by-step instructions and then rely on the action grounding to locate corresponding executable GUI elements\cite{agashe2024agent, zheng2023seeact, koh2024visualwebarena}.
While MLLMs excel in planning, their weak zero-shot grounding performance limits the effectiveness of GUI agents \cite{zheng2024gpt}. Consequently, GUI action grounding has become a focus of recent specialized model development efforts \cite{gou2024uground, cheng2024seeclick, liu2024harnessing, chen2024guicource}.

Most advanced GUI models capable of performing GUI action grounding tasks are fine-tuned from pre-trained MLLMs using substantial amounts of training data within the GUI domain. While the training data for these tasks may inevitably leave certain environments uncovered, prior research has largely relied on the generalizability of these models to transfer skills learned from the training data to novel environments. However, in real-world use cases, the action grounding task often requires environment-specific knowledge that is unique to particular settings and difficult to generalize across environments. As illustrated in the inference example at the bottom of Figure~\ref{fig:1}, the query relates to the action outcome of a triple-dot icon in the layer panel. GUI action grounding models not trained within the specific environment can hardly know that the triple-dot icon reveals layer shadow options, making it unlikely for the model to ground the query correctly. Drawing from daily experience—where prior familiarity with an environment allows humans to navigate it more effectively—we argue that when the inference environment is known to involve some novel environment, aligning GUI action grounding models to the novel environment can significantly enhance their performance.


In this work, we propose to align GUI action grounding models, originating from pre-trained MLLMs to novel GUI environments that are not trained on during the previous fine-tuning process, as shown in the top of Figure \ref{fig:1}. This method enables GUI developers to strengthen action grounding models for their specific novel use cases. The process mainly includes efficiently collecting high-quality data in the inference environments and fine-tuning the models. To collect data in any GUI environment, we introduce the GUI-Bee agent. This MLLM-based agent can autonomously explore GUI environments, where it predicts GUI actions and gathers GUI screens after each action is executed. These data from the exploration form the explorations graph, which is further transformed to data for aligning models to the environments explored. 

To optimize the efficiency and data diversity of GUI-Bee's exploration, we equip it with a novel Q-value-incentive In-context Reinforcement Learning (Q-ICRL) method. Take advantage of the in-context reasoning ability of MLLMs, Q-ICRL is training-free and relies on a memory-based way to effectively boost optimal actions during the exploration especially as more exploration actions have been made. The Q-ICRL estimates the outcomes of GUI action candidates given the current exploration status and past exploration history, helping the agent to avoid invalid or repeated actions and select actions that make the exploration cover more information.

To evaluate how GUI-Bee could help align GUI action grounding models to novel environments, we propose the NovelScreenSpot benchmark. The benchmark requires continual fine-tuning of several existing GUI grounding models to improve their performance on five novel GUI environments that they are not previously trained on. NovelScreenSpot features human-collected queries requiring rich environment-specific knowledge. 
In the experiments, we first align models to the five GUI environments by leveraging the GUI-Bee agent to explore the environments and fine-tune the  models with the collected data.
The results show that models after the alignment significantly outperform their pre-aligned counterparts, confirming the effectiveness of the collected data. 
Additionally, we perform an ablation study to evaluate the GUI-Bee agent with newly proposed metrics for screen diversity coverage and environment knowledge coverage. Our findings reveal that the Q-ICRL method boosts the efficiency of the exploration for collecting high-quality data.

The overall contributions of this paper are as follows:
\begin{itemize}
    \item We propose to align GUI grounding models to specific GUI environments required during inference, equipping them with environment-specific knowledge and boosting their performance efficiently and effectively.

    \item We introduce the GUI-Bee agent with Q-ICRL method, designed to autonomously explore GUI environments and generate high-quality data.

    \item We align GUI action grounding models to five novel environments using data collected by GUI-Bee and evaluate their performance with the NovelScreenSpot benchmark.

    \item We conduct evaluations with novel metrics for the GUI-Bee agent in generating broad and diverse data from exploration, demonstrating the effectiveness of the Q-ICRL method against baselines.

\end{itemize}

\section{Related Works}

%
%

\subsection{GUI Grounding with MLLMs}

As GUI agents evolve from text-based \cite{sodhi2024step, zhou2023webarena} to multimodal systems \cite{deng2023mindweb, koh2024visualwebarena, OSWorld}, GUI action grounding—linking natural language queries to GUI elements—has become a key challenge and shifted from being part of GUI agent tasks to an independent focus. Starting from GUI action grounding solutions \cite{zheng2024gpt, koh2024tree} that zero-shot MLLMs with the SoM methods \cite{yang2023setofmark}, recent works on GUI grounding models have put efforts into emphasizing generalization. Pioneering works like SeeClick \cite{cheng2024seeclick} introduced visual-only grounding that is easier generalized to different platforms, avoiding the limitations of platform-specific structured text, while studies such as GUICourse \cite{chen2024guicource} expanded beyond grounding executable GUI elements to general GUI element. More recent works further address challenges introduced by screenshots like high resolution and interleaved text and images by enhancing model designs \cite{gou2024uground, lin2024showui, you2024ferret} and leveraging large-scale in-domain training data \cite{wu2024atlas, liu2024harnessing}.
However, these previous works overlook that GUI action grounding queries are often environment-dependent. In our work, we boost GUI action grounding models by continuously fine-tuning them with environment-specific data collected by our GUI-Bee agent. Our method can be applied on top of all other GUI grounding models and significantly improve their performance.


\subsection{In-Context Learning }

In-context learning (ICL) refers to the method of adapting models to new tasks by providing context \cite{brown2020language, chan2022data, wang2023label}. By including examples directly in prompts, ICL allows large language models (LLMs) to generalize to unseen tasks \cite{garg2022can, pan2023context,wei2023larger}.
Prior works have explored applying ICL to reinforcement learning (RL) either with model training involved \cite{laskin2022context, lee2024supervised,xu2022prompting} or by directly leveraging pre-trained LLMs \cite{krishnamurthy2024can, monea2024llms}. We propose the Q-value-incentive In-context Reinforcement Learning (Q-ICRL), which also utilizes pre-trained LLMs but distinguishes itself by using ICL to predict state-action values. Our approach combines the adaptability of LLMs with RL's optimization-driven structure, enabling efficient action selection in the GUI environment exploration.


\begin{figure}[t]
    \centering
    \includegraphics[width=\columnwidth]{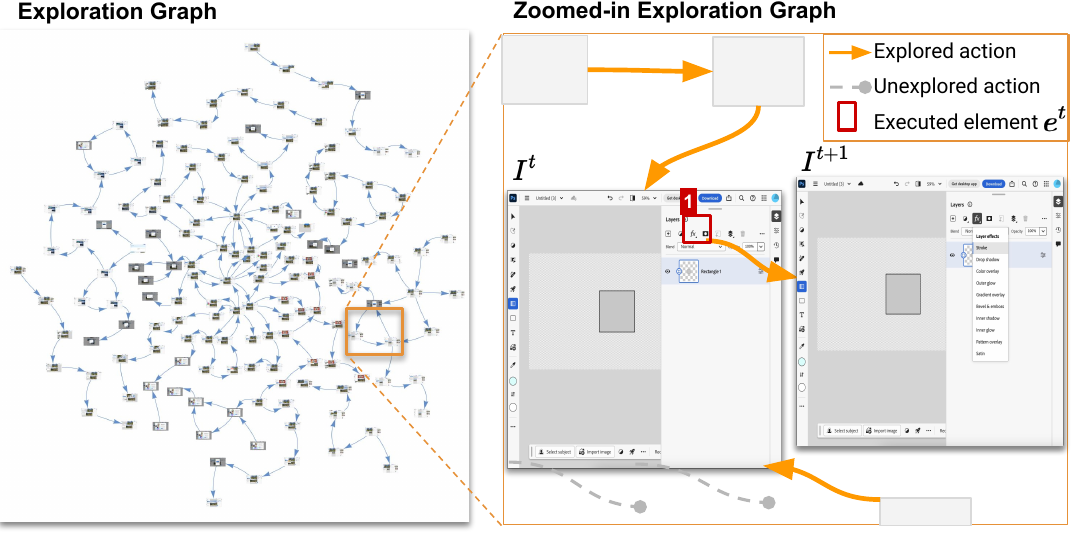}
    \caption{An example of the exploration graph showing screens connected by actions. Middle: A zoomed-in view of the graph with examples of $i^t$ and $i^{t+1}$ and some explored and unexplored actions (ever/never selected during the exploration).}
    \label{fig:2}
\end{figure}

\begin{algorithm}[t]
\fontsize{9.5}{11}\selectfont
\begin{algorithmic}
\STATE \textbf{Input:} Environment \(\text{Env}\), Initial screen \(i^1\), Maximum exploration steps \(T\)
\STATE \textbf{Output:} Exploration graph \(G\)

\STATE Initialize \(G = \{i^1\}\)

\FOR{each exploration step \(t \in [1,T]\)}
    \STATE \(S^t = G\)
    \STATE \(A_{\text{env}}(i^t) \gets \text{Env.get\_candidate\_actions}(i^t)\) 

    \STATE \(A'_{\text{env}}(i^t) \gets \textsc{WeightedSample}(A_{\text{env}}(i^t), Q)\)


    \STATE Initialize \(a^t \gets \text{None}, \text{max\_score} \gets -\infty\)
    \FOR{\(a \in A'_{\text{env}}(i^t)\)}
        \STATE \(a_{\text{eg}} \gets \textsc{ExampleActionIdentification}(a, S^t, Q)\)
        \STATE \(\text{score} \gets \textsc{InContextScoring}(a, a_{\text{eg}}, Q)\)
        \IF{\(\text{score} > \text{max\_score}\)}
            \STATE \(a^t \gets a\)
            \STATE \(\text{max\_score} \gets \text{score}\)
        \ENDIF
    \ENDFOR

    \STATE \(i^{t+1} \gets \) Env.execute(\(a^t\)) 

    \IF{\(i^{t+1} \notin G\)}
        \STATE \(G.\text{add\_node}(i^{t+1})\)
        \STATE \(G.\text{add\_edge}(i^t, a^t, i^{t+1})\)
    \ENDIF

    \STATE \textsc{UpdateQValues}\((Q, a^t, i^t, i^{t+1}, A_{\text{env}}(i^{t+1}))\)

    \STATE \(G \gets G \oplus \{a^{t}, i^{t+1}\}\)
\ENDFOR

\end{algorithmic}

\caption{Q-value-incentive In-context Reinforcement Learning (Q-ICRL)}
\label{alg:qicrl}
\end{algorithm}

\section{Aligning GUI Action Grounding Models to Novel Environments with GUI-Bee}

In this work, we focus on aligning GUI action grounding models to novel GUI environments that are previously not involved in the model training. It is a solution that enables GUI developers to build models tailored to their specific use cases. To realize, we leverage the proposed GUI-Bee agent, which autonomously collects data enriched with environment-specific knowledge through exploration and data annotation. Using this data, we continuously fine-tune the GUI action grounding models to boost their performance. The processes of exploration, data annotation, and fine-tuning are detailed in the following sub-sections.




%

\subsection{Autonomous Exploration via GUI-Bee}

\subsubsection{Exploration Goal}
\label{Exp_goal}
The goal of the exploration process is to construct an exploration graph \(G\), where GUI screens \(I\) are represented as unique nodes and GUI actions \(a\) form the edges connecting these nodes, corresponding to screen transitions. During exploration, GUI-Bee predicts actions to interact with the GUI and captures the screens before and after each action to populate the graph.
The process begins from a predefined initial GUI screen \(i^1\) at exploration step 1. At each subsequent step \(t, t \in [1, t_{max}]\), where \(t_{max}\) is the maximum number of exploration steps, the agent observes the current screen \(i^t\) and leverages an MLLM to predict the action \(a^t\). 
After executing $a^t$, if ${i}^{t+1}$ does not exist in the exploration graph $G$, it will be added to the graph as a new node, and similarly, $a^t$ will be added as a new edge if there is no existing edge in the graph between $i^{t}$ and $i^{t+1}$. An example of the exploration graph is shown in Figure \ref{fig:2}.

\begin{figure}[t]
    \centering
    \includegraphics[width=\columnwidth]{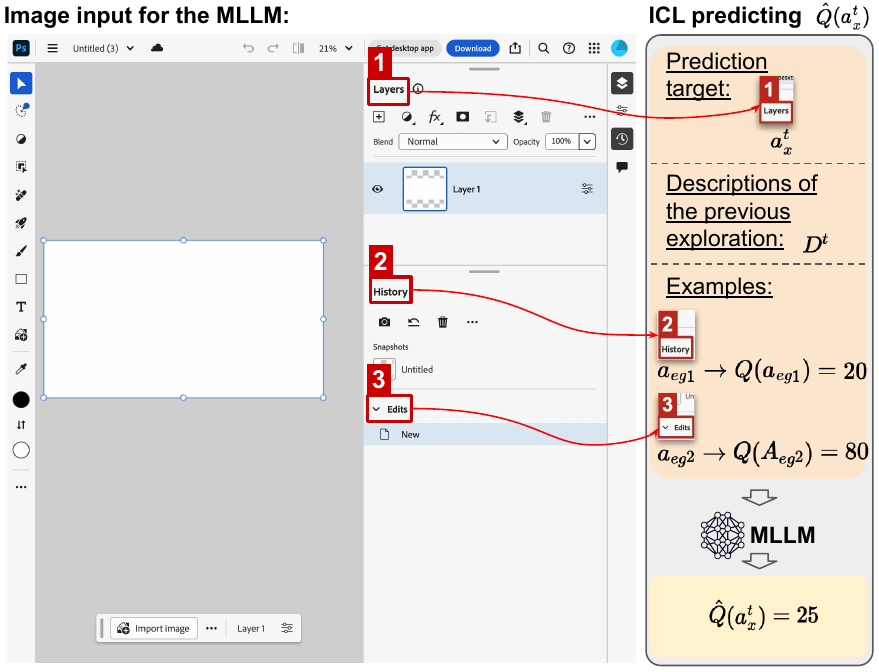}
    \caption{Example of predicting the \(\hat{Q}(a^t_x)\) with the MLLM through in-context learning (ICL). Two example actions \((a_{\text{eg1}}, a_{\text{eg2}})\) marked by bounding boxes 2 and 3 are provided as the context along with their Q values.}

    \label{fig:3}
\end{figure}

\subsubsection{Challenges in the Exploration}
\label{Exp_challenge}
The exploration process faces several challenges, starting with identifying valid actions within a noisy action space. The set of action candidates \(A_{\text{env}}(i^t)\), obtained from the environment, is defined as
\(
a^t \in 
A_{\text{env}}(i^t) = \{a^t_1, a^t_2, \dots, a^t_n\},
\)
and can originate from sources such as a Document Object Model (DOM) tree or a separate module \(M_{\text{action}}(i^t)\), like OmniParser \cite{lu2024omniparserpurevisionbased}. However, \(A_{\text{env}}(i^t)\) often includes invalid actions targeting non-executable elements, requiring the agent to discern optimal actions that are both valid and meaningful for exploration.

Another significant challenge lies in the uncertainty of screen transitions. The outcomes of executing candidate actions are unknown beforehand, and transitions are often irreversible, complicating the decision-making process. Successful exploration covers as diverse content as possible from the environment, which demands a balance between exploring new states and exploiting known beneficial actions. This balance necessitates accurate action prediction and robust reasoning, making it critical for the agent to effectively navigate the action space and handle the inherent complexities of GUI environments.

\subsubsection{Q-value-Incentive In-Context Reinforcement Learning (Q-ICRL)}
\label{qicrl}
We consider the exploration process as a Markov Decision Problem, which is defined by a tuple \(\langle \mathcal{S}, \mathcal{A}, P, r \rangle\), where \(\mathcal{S}\) denotes the state space, \(\mathcal{A}\) represents the action space, \(P : \mathcal{S} \times \mathcal{A} \times \mathcal{S} \to \{0, 1\}\) is the state transition probability function, and \(r : \mathcal{S} \times \mathcal{A} \to \mathbb{R}\) denotes the reward function. At each exploration step \(t \in \mathbb{N}\), the GUI-Bee agent is at \(S^t \in \mathcal{S}\), and takes an action \(a^t \in \mathcal{A}\) on the current observed screen \(i^t\) which transitions to a new state \(S^{t+1} \in \mathcal{S}\) with probability \(P(S^{t+1} | S^t, a^t)\), receiving a reward \(r(S^t, a^t)\). The reward \(r\) is binary, where it is positive when the \(a^t\) leads to a new screen not existing in the exploration graph at the beginning of the current exploration step, i.e., \(i^{t+1} \notin G^{t-1}\). Accordingly, the state \(S^t\) is defined as the exploration graph \(G^{t-1}\) to satisfy the Markov property, where \(S^1\) contains only the initial screen. For simplicity, \(S^t\) is approximated by a set of natural language descriptions \(D^t = \{d^k \mid d^k = \text{Describe}(a^k, G^{t-1}), a^k \in G^{t-1}\}\), where \(a^k\) represents the edges in the exploration graph and \(\text{Describe}(\cdot)\) uses an MLLM to generate descriptions for the actions and screens before and after the action. Further details on this process are provided in Appendix \ref{Appendix_approx_s_with_d}.

We propose the Q-value-Incentive In-Context Reinforcement Learning (Q-ICRL) method, outlined in Algorithm \ref{alg:qicrl}, to maximize the accumulated rewards in the exploration. Q-ICRL quantifies the future rewards of executing actions \(a^t\) at \(S^t\) with a Q-value function \(Q(S^t, a^t)\). Unlike traditional Q-learning, Q-ICRL is a training-free algorithm as the \(Q\) is mainly memory-based, which is detailed in the following subsection.
To select an action \(a^t\) at state \(S^t\), Q-values \(Q(S^t, a^t_i)\) are first used as weights to sample a subset \(A'_{\text{env}}(i^t)\) of length \(H\) from the action space \(A_{\text{env}}(i^t)\). Then, an MLLM with in-context learning is employed to identify the most promising action \(a^t\) from \(A'_{\text{env}}(i^t)\), guided by the Q-value function. This process involves two steps: \(\textsc{ExampleActionIdentification}\) and \(\textsc{InContextScoring}\), both detailed in the next subsection. This method ensures a balance between exploration and exploitation.
Once an action \(a^t\) is selected, it is executed in the environment, leading to an updated exploration graph \(G^{t+1} = G^t \oplus (a^t, i^{t+1})\), where \(\oplus\) denotes adding a new node and edge to the graph. The Q-value function is updated accordingly, as described in Section \ref{Exp_goal}.

\paragraph{Q-value Function}
\label{qIU}

Given $a^t_i$ in \(A_{\text{env}}(i^t)\), the \(Q(S^t, a^t_i)\) should be relatively low if executing it leads to a $i^{t+1}$ that is repeated in \(S^t\), however, as mentioned in Section \ref{Exp_challenge}, the $i^{t+1}$ unknown at \(S^t\). To overcome this problem, We propose a Q-value function that approximates \(Q(S^t, a^t_i)\) with \(Q(S^x, a^t_i)\) when there is a \(S^x\), \(x \in [1, \dots, t-1]\) with the same \(a^t_i\) was executed. As for \(a^t_i\) that is never executed in the past states, \(Q(S^t, a^t_i)\) is set at a default value of 100.


After executing \(a^t_i\), the \(Q(S^t, a^t)\) values are updated based on the outcome of execution, reflecting the desirability of the action's result:
\[
    Q(S^t, a^t)  = \begin{cases}
\gamma_{\text{max}} \cdot Q_{Next} \text{, if } I_{t+1} \text{ is an unseen screen,} \\
\gamma_{\text{med}}\cdot Q_{Next} \text{, if } I_{t+1} \text{ is a seen screen,}\\
\gamma_{\text{low}}\cdot Q(S^t, a^t) \text{, }\text{ if } I_{t+1} \text{ is the same as }I_{t}.
\end{cases}
\\
\]

Here, \(\gamma_{\text{max}} > \gamma_{\text{med}} > \gamma_{\text{low}}\), which are hyper parameters that we set to be $0.85$, $0.75$ and $0.4$ respectively. \(Q_{Next}\) represents the average \(Q\)-value of all candidate actions in the next state \(S^{t+1}\), computed as:
\(
Q_{Next} = \text{Mean}(\{Q(S^{t+1}, a^{t+1}_i) \mid a^{t+1}_i \in A_{\text{env}}(i^{t+1})\}).
\)
This update mechanism ensures that \(Q(S^t, a^t)\) is dynamically adjusted based on the action's outcome. Actions leading to new screens with more unseen candidate actions are rewarded with higher values, while actions leading to redundant or ineffective transitions are penalized. Additionally, the \(Q\)-values for valid but repeatedly executed actions gradually decrease, promoting exploration of less frequently chosen alternatives.

\paragraph{In-context Action Selection}
\label{incontextAS}

To offset the potential error in the Q-value function, we leverage the knowledge in the MLLM to determine the most suitable action. After \(A'_{\text{env}}(i^t)\) is sampled, we let the MLLM to predict \(\hat{Q}(S^t, a^t_x)\) for each sampled candidate action \(a^t_x \in A'_{\text{env}}(i^t)\). \(\hat{Q}(a^t_x)\) is an MLLM-approximation of the \(Q(S^t, a^t_x)\), and the in-context learning is employed. Specifically, the MLLM is provided with the current state \(S^t\) and two example actions \((a_{\text{eg1}}, a_{\text{eg2}})\) along with their corresponding Q-values \((Q(S^t, a_{\text{eg1}}), Q(S^t, a_{\text{eg2}}))\).
The example actions are chosen from \(A_{\text{env}}(i^t)\) with \((Q(S^t, a_{\text{eg1}}), Q(S^t, a_{\text{eg2}})) \neq 100\) while having the corresponding element \(e_{\text{eg1}}\) and \(e_{\text{eg2}}\) most similar to the \(e^t_x\). We develop a GUI Element Fuzzy Visual Matching module, detailed in Appendix \ref{Appendix_fuzz_matching}, to identify the visual similarity between element visuals. Next, we mark \((a_{\text{eg1}}, a_{\text{eg2}}) \text{ and } a^t_x\) on the screenshot of $i^t$ as the visual input to the MLLM with the Set-of-Mark method \cite{yang2023setofmark}. Figure \ref{fig:3} shows an example of this process, and the full prompts are detailed in Appendix \ref{Appendix_ICL}. If no suitable  \(a_{\text{eg}}\) is found, the example-related content is omitted from the prompt. Finally, the action \(a^t_x\) with the highest predicted \(\hat{Q}(a^t_x)\) is selected as the action \(a^t\) to be executed.

\subsection{Autonomous Data Annotation with GUI-Bee}

After the exploration, for each edge $a^t$ in the $G$, the connected nodes $(i^t, i^{t+1})$ is sent to the MLLM to generate $u^t$, a list of queries serving as action grounding queries for the target element $e^t$ in ${i}^t$. We carefully design the prompt, detailed in Appendix \ref{Appendix_prompt_for_u}, to guide the MLLM in generating these queries. The queries involve both "what is currently visible" on the screen and "what will appear" after interacting with GUI elements, which we refer as System 1 and System 2 grounding queries, inspired by \citet{kahneman2011thinking}. Building on the multi-lens prompting method \cite{fan2024read}, we create separate visual prompts, or lenses: two capture the full screens of $i^t$ and $i^{t+1}$ while one isolates $e^t$ to ensure high-quality outputs. The center of $e^t$ is then sampled as the target point $p^t$, and the resulting data ($u^t, {i}^t, p^t$) is used to fine-tune GUI action grounding models. 

\subsection{Fine-tuning Models with Environment-specific Data}
We leverage the data generated by the GUI-Bee agent's exploration and annotation processes to fine-tune GUI grounding models to align them with specific novel GUI environments. The data consists of pairs of inputs, including the GUI screen $I$ (comprising a screenshot and an accessibility tree, detailed in Appendix \ref{appendix_exp_d}) and grounding query $u$, along with corresponding outputs, the target location $p$. 
Benefiting from the flexibility of the representation of $I$ in the generated exploration data, we are able to fine-tune models in two input configurations: vision-only and Vision+A11y. In the vision-only configuration, the input consists solely of GUI screenshots. In the Vision+A11y configuration, the input includes both GUI screenshots and the accessibility tree (A11y tree), embedded as part of the text prompt. The fine-tuning process aims to adapt the models to leverage environment-specific knowledge efficiently, improving their grounding performance in the target environments.

\begin{table*}[t]
    \centering
\resizebox{\textwidth}{!}{
\begin{tabular}{llccccc}
                                                                                                       &                                     & \multicolumn{1}{l}{} & \multicolumn{1}{l}{} & \multicolumn{1}{l}{} & \multicolumn{1}{l}{} & \multicolumn{1}{l}{} \\ \hline
\multicolumn{1}{l}{}                                                                                  &                                     & \textbf{Shopping}    & \textbf{Classifieds} & \textbf{Reddit}    & \textbf{Eventbrite}  & \textbf{Photoshop-web}   \\ \midrule
\multicolumn{1}{l|}{\multirow{3}{*}{\begin{tabular}[c]{@{}l@{}}NovelScreenSpot\\ Benchmark\end{tabular}}} & \# Unique Action Grounding Targets  & 58                   & 42                   & 44                   & 47                   & 44                   \\
\multicolumn{1}{l|}{}                                                                                  & \# Unique Grounding Queries         & 105                  & 98                   & 96                   & 107                  & 106                  \\
\multicolumn{1}{l|}{}                                                                                  & Ratio of Queries about Action Outcomes   & 30.5\%               & 32.7\%               & 39.6\%               & 42.1\%               & 34.0\%               \\
\midrule
\multicolumn{1}{l|}{\multirow{2}{*}{\begin{tabular}[c]{@{}l@{}}Exploration \\ Generated Data\end{tabular}}} & \# Unique Action Grounding Targets  & 555                  & 530                  & 590                  & 526                  & 692                  \\
\multicolumn{1}{l|}{}                                                                                  & \# Unique Grounding Queries         & 6,080                & 5,719                & 6,480                & 5,740                & 6,876               \\
\bottomrule
\end{tabular}}
\caption{Statistics of the NovelScreenSpot benchmark data and the exploration data generated by GUI-Bee. The generated data is used to continuously fine-tune GUI grounding models which are then evaluated by the NovelScreenSpot benchmark.}
\label{table1}

\end{table*}

\section{NovelScreenSpot Benchmark} 
We propose NovelScreenSpot, a benchmark for evaluating GUI action grounding models in five diverse web GUI environments and their performance improvements after they are continuously fine-tuned with new data. It includes triplets of queries, screens represented by screenshots and accessibility trees, and ground-truth bounding boxes for the grounding targets. We show the statistics of the NovelScreenSpot in Table \ref{table1} and examples in Appendix \ref{appendix_NSS_example}.

NovelScreenSpot simulates real-world GUI model deployment scenarios, where App owners want to evaluate GUI action models on their own specific GUI environments, potentially novel environments that the models are not previously trained on.
Therefore, unlike existing benchmarks that emphasize diversity across many environments, such as the ScreenSpot \cite{cheng2024seeclick}, NovelScreenSpot instead provides a greater data variation within each environment. Notably, the NovelScreenSpot includes a large number of queries, around one-third of the total benchmark, focusing on interaction outcomes. These queries require environment-specific knowledge and hardly exist in the existing benchmarks.
The five GUI environments in the NovelScreenSpot are three offline websites from the VisualWebArena \cite{koh2024visualwebarena}—Shopping, Classifieds (a second-hand marketplace), and Reddit (an online forum)—and two online websites, Photoshop-web and Eventbrite. These environments vary greatly in style, with Photoshop-web dominated by professional icons, Shopping and Classifieds emphasizing images, and Reddit and Eventbrite focusing more on textual content.

\paragraph{Task and Metrics} The models are required to predict points within the target GUI elements corresponding to the language queries in NovelScreenSpot, simulating how users indicate a GUI element with a cursor. We test each GUI action grounding model on the NovelEneSpot twice, before and after continual fine-tuning with new data. By comparing the difference between the two benchmark results, we quantify their performance improvements on each GUI environment in the NovelScreenSpot.
We define two testing scenarios: \textit{vision+A11y}, where the model input includes the GUI screenshot, query, and a text string containing the accessibility tree, and \textit{vision-only}, where the input is only the screenshot and query. The predicted point is considered correct if it is inside the ground truth-bounding box of the target element, and the model performance is evaluated using the accuracy of predicted points.

\paragraph{Annotation}
NovelScreenSpot is manually constructed through a multi-step annotation process to ensure high-quality and unambiguous data. First, we ask annotators to interact with the web environments and record their actions, including the screens before and after the action and the corresponding target elements. In the second step, different annotators write queries corresponding to these recorded actions. To guide query creation, we provide annotators with hints to focus on three perspectives: (1) the direct name or label of the target element, (2) the appearance of the element, and (3) the outcome of interacting with the element.  
Lastly, we manually validate the annotation results to ensure quality. This includes eliminating ambiguous queries, removing duplicates, and discarding low-quality annotations that do not align with the recorded actions. The final dataset provides clear triplets of queries, screens (represented by screenshots and accessibility trees), and target elements corresponding to the queries. An example of the annotation interface used for collecting GUI action grounding queries is shown in Appendix \ref{appendix-annotation}.

\section{Experiments}



{\renewcommand{\arraystretch}{0.8}

\begin{table*}[tbp]
    \centering
      \setlength\tabcolsep{6pt}

\resizebox{\textwidth}{!}{
\begin{tabular}{l l l l l l l | l}
& \multicolumn{6}{c}{\textbf{ \small NovelScreenSpot}} & \textbf{ \small Multimodal-M2W}\\
\toprule
& \textbf{Shopping} & \textbf{Classifieds} & \textbf{Reddit} & \textbf{Eventbrite} & \textbf{Photoshop-web} & \textbf{Avg.} &  \textbf{Eventbrite}  \\ \midrule
\multicolumn{8}{c}{\textit{Vision-only GUI Action Grounding}}    \\ \midrule
SeeClick & 36.2 & 36.7 & 35.4 & 35.5 & 13.2 & - &  23.1  \\
$\text{SeeClick}_{\text{Mind2Web}}$
 & 39.0 {\footnotesize\textit{(+2.8)}} 
 & 30.6 {\footnotesize\textit{(-6.1)}} 
 & 36.5 {\footnotesize\textit{(+1.1)}} 
 & 43.0 {\footnotesize\textit{(+7.5)}} 
 & 9.4 {\footnotesize\textit{(-3.8)}} 
 & {\footnotesize\textit{(+0.3)}} & 38.5 {\footnotesize\textit{(+15.4)}} \\
$\text{SeeClick}_{\text{GUI-Bee}}$(Ours)
 & 48.6 {\footnotesize\textit{(+12.4)}} 
 & 44.9 {\footnotesize\textit{(+8.2)}} 
 & 39.6 {\footnotesize\textit{(+4.2)}} 
 & 53.3 {\footnotesize\textit{(+17.8)}} 
 & 18.9 {\footnotesize\textit{(+5.7)}} 
 & {\footnotesize\textit{(+9.7)}} &    38.5 {\footnotesize\textit{(+15.4)}} \\[6pt]

UIX-7B & 31.4 & 38.8 & 44.8 & 43.0 & 22.6 & - & 23.1  \\
$\text{UIX-7B}_{\text{Mind2Web}}$
 & 39.0 {\footnotesize\textit{(+7.6)}} 
 & 38.8 {\footnotesize\textit{(+0)}} 
 & 37.5 {\footnotesize\textit{(+7.3)}} 
 & 43.9 {\footnotesize\textit{(+0.9)}} 
 & 16.0 {\footnotesize\textit{(-6.6)}} 
 & {\footnotesize\textit{(+1.8)}} &  23.1 {\footnotesize\textit{(+0)}} \\
 
$\text{UIX-7B}_{\text{GUI-Bee}}$(Ours)
 & 78.1 {\footnotesize\textit{(+46.7)}} 
 & 66.3 {\footnotesize\textit{(+27.5)}} 
 & 60.4 {\footnotesize\textit{(+15.6)}} 
 & 70.1 {\footnotesize\textit{(+27.1)}} 
 & 31.1 {\footnotesize\textit{(+8.5)}} 
 & {\footnotesize\textit{(+25.1)}} &  53.8 {\footnotesize\textit{(+30.7)}} \\[6pt]
 
Qwen-GUI & 19.0 & 21.4 & 26.0 & 34.6 & 9.4 & - & 23.1 \\
$\text{Qwen-GUI}_{\text{Mind2Web}}$
 & 23.1 {\footnotesize\textit{(+4.1)}} 
 & 21.6 {\footnotesize\textit{(+0.2)}} 
 & 28.1 {\footnotesize\textit{(+2.1)}} 
 & 36.0 {\footnotesize\textit{(+1.4)}} 
 & 10.3 {\footnotesize\textit{(+0.9)}} 
 & {\footnotesize\textit{(+1.7)}} & 23.1 {\footnotesize\textit{(+0)}}\\
 $\text{Qwen-GUI}_{\text{GUI-Bee}}$(Ours)
 & 25.7 {\footnotesize\textit{(+6.7)}} 
 & 31.6 {\footnotesize\textit{(+10.2)}} 
 & 28.1 {\footnotesize\textit{(+2.1)}} 
 & 37.4 {\footnotesize\textit{(+2.8)}} 
 & 12.3 {\footnotesize\textit{(+2.9)}} 
 & {\footnotesize\textit{(+4.9)}} &  \textbf{46.2} {\footnotesize\textit{(+23.1)}}  \\[6pt]


\midrule
\multicolumn{8}{c}{\textit{Vision+A11y GUI Action Grounding}} \\ \midrule
Qwen-GUI & 34.3 & 50.0 & 34.4 & 52.3 & 13.2 & - & -  \\
$\text{Qwen-GUI}_{\text{GUI-Bee}}$(Ours)
 & 51.4 {\footnotesize\textit{(+17.1)}} 
 & 54.1 {\footnotesize\textit{(+4.1)}} 
 & 55.2 {\footnotesize\textit{(+20.8)}} 
 & 62.6 {\footnotesize\textit{(+10.3)}} 
 & 41.5 {\footnotesize\textit{(+28.3)}} 
 & {\footnotesize\textit{(+16.1)}} &  -  \\[6pt]

UIX-7B & 16.2 & 14.3 & 11.5 & 21.5 & 10.4 & - & - \\
$\text{UIX-7B}_{\text{GUI-Bee}}$(Ours)
 & \textbf{74.3} {\footnotesize\textit{(+58.1)}} 
 & \textbf{77.6} {\footnotesize\textit{(+63.3)}} 
 & \textbf{80.2} {\footnotesize\textit{(+68.7)}} 
 & \textbf{82.2} {\footnotesize\textit{(+60.7)}} 
 & \textbf{70.8} {\footnotesize\textit{(+60.4)}} 
 & {\footnotesize\textit{(+62.2)}} &   -  \\[6pt]

\bottomrule
\end{tabular}}
\caption{Results of benchmarking GUI grounding models on the NovelScreenSpot and Eventbrite environment of Multimodal-Mind2Web benchmark. We show the model accuracy and, in parentheses, the absolute improvement over the vanilla models after the models are continuously fine-tuned. The results demonstrate that our GUI-Bee model significantly improves the performance of GUI action grounding models in novel environments.}
\label{tab2}
\end{table*}
}





\subsection{Alignment to Novel GUI Environments}
In the first experiment, we leverage the NovelScreenSpot benchmark to evaluate how GUI-Bee collected data help align GUI action grounding models to novel environments. We first select several GUI action grounding models and identify the GUI environments within the NovelScreenSpot that are novel for them. Then we attempt to align the models to their corresponding novel environments.

\paragraph{Setups}
We employ the GUI-Bee agent to explore the five environments in NovelScreenSpot, conducting up to 400 exploration steps per environment with three candidate actions sampled at each step. To ensure diverse screen data, the exploration is repeated three times per environment at varying screen resolutions. In this work, we adopt GPT-4o \cite{gpt4o} as the multimodal large language model (MLLM), though other MLLMs can also be integrated into the framework. The resulting exploration statistics are summarized in Table \ref{table1}.
Using the data generated from these explorations, we continuously fine-tune four GUI grounding models: SeeClick \cite{cheng2024seeclick}, Qwen-GUI \cite{chen2024guicource}, and UIX-7B \cite{liu2024harnessing} 
to adapt to novel environments within NovelScreenSpot. 
Additionally, for the alignment of Qwen-GUI and UIX-7B, we leverage both vision-only and Vision+A11y data, as these models were previously fine-tuned to accommodate similar formats. Further details on exploration configurations, data formatting, and fine-tuning settings are provided in Appendix \ref{appendix_exp_d}.

\paragraph{Main Results}
We present the model performance before and after alignment to novel GUI environments in Table \ref{tab2}. The models are either continuously fine-tuned using exploration data collected by the GUI-Bee agent or, as a baseline, training data from Multimodal-Mind2Web \cite{zheng2023seeact}. Our results show that data collected by GUI-Bee significantly outperforms the baseline in improving model performance across all tested models. We attribute this to the environment-specific knowledge captured by the GUI-Bee agent, which contrasts with Multimodal-Mind2Web's design for generalization across environments. Notably, the UIX-7B model achieves the most substantial accuracy gains and the highest overall accuracy after alignment. 
Furthermore, models fine-tuned in the \textit{vision+A11y} GUI action grounding setting demonstrate greater improvements. This is likely because the original GUI grounding models were not trained with accessibility (A11y) tree information, and the inclusion of A11y data provides additional screen context, enabling more accurate predictions. These findings underscore the value of the diverse formats in GUI-Bee's collected data, which incorporate contextual information such as accessibility data to enhance model performance.

\paragraph{Results on Grounding for Offline GUI Agents}
We extend our evaluation by following \citet{gou2024uground} to leverage the test data from Multimodal-Mind2Web \cite{zheng2023seeact} for assessing the alignment of GUI grounding models to novel environments. This test data simulates GUI grounding tasks guided by MLLM-planned instructions in GUI agent applications. Multimodal-Mind2Web data is derived from real action sequences across diverse GUI environments, corresponding to high-level, real-world GUI tasks. The grounding queries consist of element descriptions generated by MLLMs, without explicitly referencing coordinates.
As shown in Table \ref{tab2}, the result reveals that models continuously fine-tuned with GUI-Bee data achieve significantly greater performance improvements compared to baseline models using Multimodal-Mind2Web training set data. This further validates the effectiveness of our proposed method.

\begin{figure}[t]
\includegraphics[width=\linewidth]{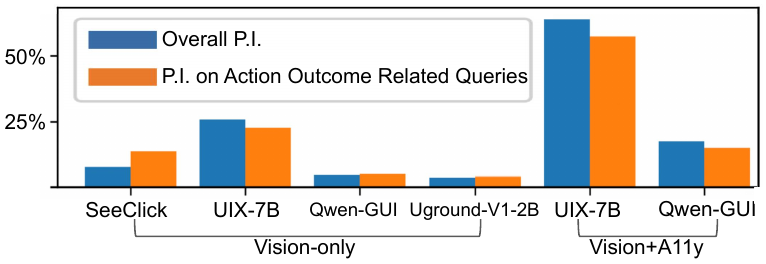}
\caption{Overall average model performance improvements (P.I.) in the NovelScreenSpot benchmark, compared with the average model P.I. on queries related to action outcome. The consistent P.I. between these two categories shows the proposed alignment improves model performance evenly.}

\label{fig:4}

\end{figure}

\paragraph{Model Performance on Action Outcomes Related Queries}    

In Figure \ref{fig:4}, we present the average performance improvements of the models on queries related to action outcomes, as well as their overall performance improvements after alignment using GUI-Bee collected data. The results show consistent gains in both overall model performance and performance on action-outcome-related queries across all models. This demonstrates that the data generated by our GUI-Bee agent is universally effective in enhancing model performance for grounding tasks in the novel GUI environment, addressing not only environment-specific challenges but also general grounding weaknesses faced by models.

\subsection{Exploration Efficiency Evaluation}
To further evaluate the efficiency of our GUI-Bee agent in exploring and generating data within GUI environments, we task it, along with two baseline agents, to explore three offline GUI environments: Shopping, Classifieds, and Reddit—from the Visual Web Arena \cite{koh2024visualwebarena}. These environments are reset to identical initial states at the beginning of each exploration, ensuring that all agents start from the same conditions and face equivalent challenges.

\paragraph{Evaluation and Metrics}

To evaluate exploration efficiency, we assess the diversity of actions and screens in the exploration graph \(G^{t_{max}}\) generated by each agent under the same maximum number of exploration steps \(t_{max}\). 

First, we convert each exploration graph into \(D^{t_{max}+1}\), the natural language approximated RL state as described in Section \ref{qicrl}. GPT-4o is then used to compare pairs \((D^{t_{max}+1}_1, D^{t_{max}+1}_2)\) from two agents at a time to determine which demonstrates broader coverage. This process, referred to as the Relative Exploration Coverage Ranking, provides an intuitive comparative measure of exploration breadth.

Further, we introduce the Depth-fixed DOM Diversity Counts (D3C) metric to assess structural variation in the screens within the exploration graph \(G^{t}\) generated by different agents objectively. D3C is defined as the number of distinct page structures in the \(G^{t}\). Each page structure is determined by truncating the DOM tree of a screen to a fixed depth, retaining only the class attributes of elements. By counting the unique page structures within all the page structures within \(G^{t}\), we get the D3C value at the exploration step \(t\). With a fixed number of exploration steps, D3C provides a quantitative measure of the agent's efficiency in uncovering diverse structural layouts, offering a clear and objective metric for exploration breadth.

\paragraph{Baselines}

To evaluate the Q-ICRL method adopted by our GUI-Bee agent, we introduce two ablated version of GUI-Bee: one explore with the In-Context Reinforcement Learning (ICRL) method and the other one explore with the random strategy. 
The ICRL agent is an ablated version of our GUI-Bee agent. It uses the same RL state representation as the GUI-Bee agent but does not build or utilize the Q-value function. Instead, it directly leverages the MLLM with in-context learning to select the next action on the GUI screen. The in-context example is randomly chosen from the actions recorded in the exploration graph.
The random agent follows a purely stochastic strategy, selecting actions randomly from the set of candidate actions on each screen. All agents are constrained to the same maximum number of action steps to enable a fair comparison.

\begin{figure}[t]
\includegraphics[width=\linewidth]{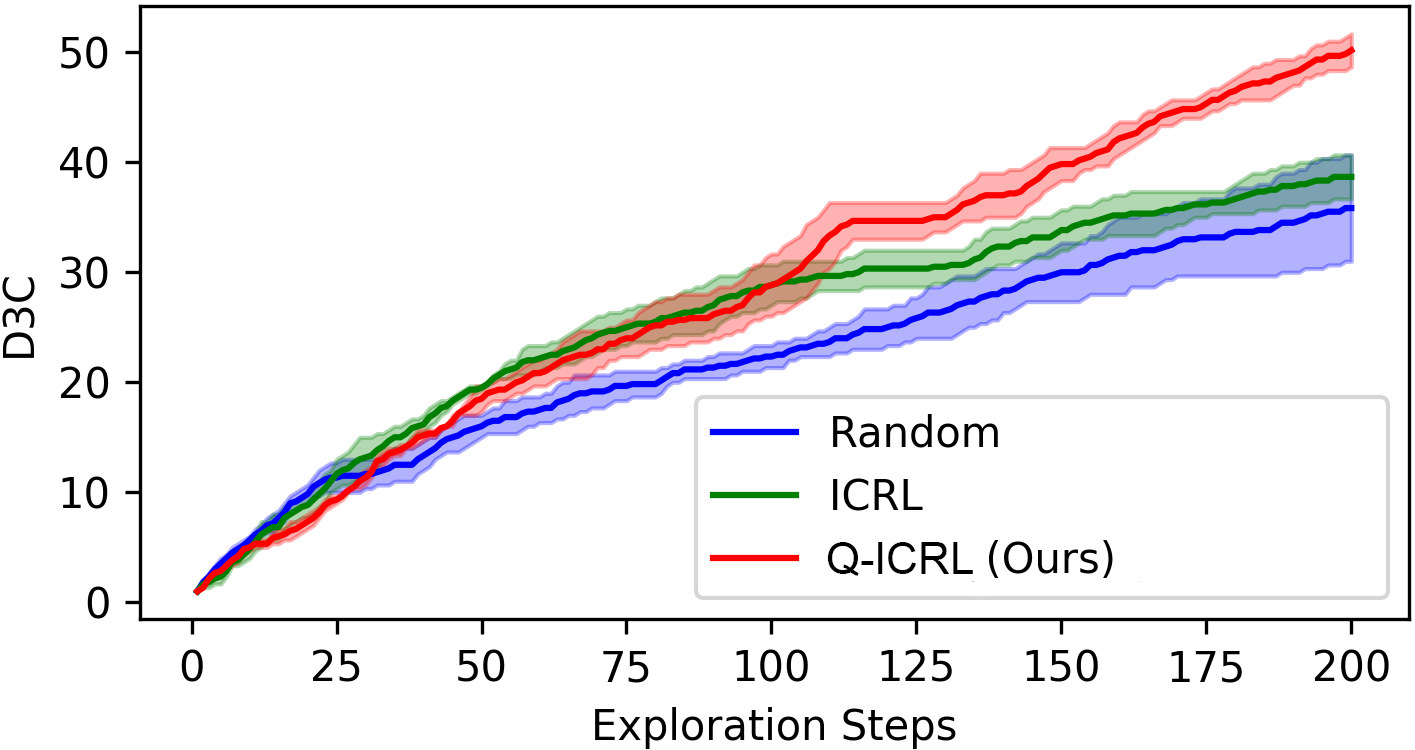}
\caption{Mean and standard deviation of Depth-fixed DOM Diversity Counts (D3C) at various exploration steps across three runs in three environments. GUI-Bee agent demonstrates a wider exploration coverage.}

\label{fig:e2_result}

\end{figure}

\paragraph{Results}
Using the exploration graphs generated by each agent, we first perform the Relative Exploration Coverage Ranking by comparing the natural language descriptions \(D^{T+1}\) of the exploration graphs produced at the end of their explorations. GPT-4o consistently identifies the \(D^{T+1}\) from our GUI-Bee agent as demonstrating broader coverage compared to the ones from the baseline agents. 
Then, we calculate the D3C for each exploration conducted by agents in all three environments and compute the average D3C for each agent across the three environments. This process is repeated three times, and the mean and standard deviation of the averaged D3C across these evaluations are plotted in Figure \ref{fig:e2_result}. As the number of exploration steps increases, the averaged D3C for all agents grows, but our GUI-Bee agent demonstrates stronger growth momentum and significantly outperforms the baseline agents after 100 exploration steps. These results collectively demonstrate that the GUI-Bee agent is more efficient in covering broader areas of the GUI environment and uncovering a more diverse range of structural layouts compared to the baseline agents.


\section{Conclusion}
This work introduces the GUI-Bee agent, a novel MLLM-based approach for autonomously exploring GUI environments and collecting high-quality, environment-specific data. We continuously fine-tune GUI action grounding models to align to novel environments using this data. Through experiments, we demonstrate significant improvements in model performance after the alignment, validating the effectiveness of our approach. Additionally, we propose novel metrics to evaluate GUI-Bee against baselines in generating diverse exploration data, further highlighting the efficiency and effectiveness of the Q-ICRL method.

\section{Limitations}

The GUI-Bee agent excels in tailoring GUI action grounding models for specific environments but has limitations. It uses Multimodal Large Language Models (MLLMs) like GPT-4o, which provide high-quality data but come with higher computational costs, latency, and privacy issues. Additionally, as exploration within a specific environment continues and the number of exploration steps increases, the GUI-Bee agent accumulates an increasingly long history of exploration data. Processing this longer history can introduce additional overhead, increasing the time required to select actions for subsequent exploration steps. This may impact scalability in highly complex environments or during extended exploration sessions.

\bibliography{custom}

\begin{thebibliography}{35}
\providecommand{\natexlab}[1]{#1}

\bibitem[{Agashe et~al.(2024)Agashe, Han, Gan, Yang, Li, and Wang}]{agashe2024agent}
Saaket Agashe, Jiuzhou Han, Shuyu Gan, Jiachen Yang, Ang Li, and Xin~Eric Wang. 2024.
\newblock Agent s: An open agentic framework that uses computers like a human.
\newblock \emph{arXiv preprint arXiv:2410.08164}.

\bibitem[{Bai et~al.(2023)Bai, Bai, Yang, Wang, Tan, Wang, Lin, Zhou, and Zhou}]{bai2023qwen}
Jinze Bai, Shuai Bai, Shusheng Yang, Shijie Wang, Sinan Tan, Peng Wang, Junyang Lin, Chang Zhou, and Jingren Zhou. 2023.
\newblock Qwen-vl: A frontier large vision-language model with versatile abilities.
\newblock \emph{arXiv preprint arXiv:2308.12966}.

\bibitem[{Brown(2020)}]{brown2020language}
Tom~B Brown. 2020.
\newblock Language models are few-shot learners.
\newblock \emph{arXiv preprint arXiv:2005.14165}.

\bibitem[{Chan et~al.(2022)Chan, Santoro, Lampinen, Wang, Singh, Richemond, McClelland, and Hill}]{chan2022data}
Stephanie Chan, Adam Santoro, Andrew Lampinen, Jane Wang, Aaditya Singh, Pierre Richemond, James McClelland, and Felix Hill. 2022.
\newblock Data distributional properties drive emergent in-context learning in transformers.
\newblock \emph{Advances in Neural Information Processing Systems}, 35:18878--18891.

\bibitem[{Chen et~al.(2024)Chen, Cui, Hu, Qin, Fang, Zhao, Wang, Liu, Chen, Huo, Yao, Lin, Liu, and Sun}]{chen2024guicource}
Wentong Chen, Junbo Cui, Jinyi Hu, Yujia Qin, Junjie Fang, Yue Zhao, Chongyi Wang, Jun Liu, Guirong Chen, Yupeng Huo, Yuan Yao, Yankai Lin, Zhiyuan Liu, and Maosong Sun. 2024.
\newblock Guicourse: From general vision language models to versatile gui agents.

\bibitem[{Cheng et~al.(2024)Cheng, Sun, Chu, Xu, Li, Zhang, and Wu}]{cheng2024seeclick}
Kanzhi Cheng, Qiushi Sun, Yougang Chu, Fangzhi Xu, Yantao Li, Jianbing Zhang, and Zhiyong Wu. 2024.
\newblock Seeclick: Harnessing gui grounding for advanced visual gui agents.
\newblock \emph{arXiv preprint arXiv:2401.10935}.

\bibitem[{Deng et~al.(2023)Deng, Gu, Zheng, Chen, Stevens, Wang, Sun, and Su}]{deng2023mindweb}
Xiang Deng, Yu~Gu, Boyuan Zheng, Shijie Chen, Samuel Stevens, Boshi Wang, Huan Sun, and Yu~Su. 2023.
\newblock \href {https://openreview.net/forum?id=kiYqbO3wqw} {Mind2web: Towards a generalist agent for the web}.
\newblock In \emph{Thirty-seventh Conference on Neural Information Processing Systems}.

\bibitem[{Fan et~al.(2024)Fan, Ding, Kuo, Jiang, Zhao, Guan, Yang, Zhang, and Wang}]{fan2024read}
Yue Fan, Lei Ding, Ching-Chen Kuo, Shan Jiang, Yang Zhao, Xinze Guan, Jie Yang, Yi~Zhang, and Xin Wang. 2024.
\newblock Read anywhere pointed: Layout-aware gui screen reading with tree-of-lens grounding.
\newblock In \emph{Proceedings of the 2024 Conference on Empirical Methods in Natural Language Processing}, pages 9503--9522.

\bibitem[{Garg et~al.(2022)Garg, Tsipras, Liang, and Valiant}]{garg2022can}
Shivam Garg, Dimitris Tsipras, Percy~S Liang, and Gregory Valiant. 2022.
\newblock What can transformers learn in-context? a case study of simple function classes.
\newblock \emph{Advances in Neural Information Processing Systems}, 35:30583--30598.

\bibitem[{Gou et~al.(2024)Gou, Wang, Zheng, Xie, Chang, Shu, Sun, and Su}]{gou2024uground}
Boyu Gou, Ruohan Wang, Boyuan Zheng, Yanan Xie, Cheng Chang, Yiheng Shu, Huan Sun, and Yu~Su. 2024.
\newblock \href {https://arxiv.org/abs/2410.05243} {Navigating the digital world as humans do: Universal visual grounding for gui agents}.
\newblock \emph{arXiv preprint arXiv:2410.05243}.

\bibitem[{Kahneman(2011)}]{kahneman2011thinking}
Daniel Kahneman. 2011.
\newblock Thinking, fast and slow.
\newblock \emph{Farrar, Straus and Giroux}.

\bibitem[{Koh et~al.(2024{\natexlab{a}})Koh, Lo, Jang, Duvvur, Lim, Huang, Neubig, Zhou, Salakhutdinov, and Fried}]{koh2024visualwebarena}
Jing~Yu Koh, Robert Lo, Lawrence Jang, Vikram Duvvur, Ming~Chong Lim, Po-Yu Huang, Graham Neubig, Shuyan Zhou, Ruslan Salakhutdinov, and Daniel Fried. 2024{\natexlab{a}}.
\newblock Visualwebarena: Evaluating multimodal agents on realistic visual web tasks.
\newblock \emph{arXiv preprint arXiv:2401.13649}.

\bibitem[{Koh et~al.(2024{\natexlab{b}})Koh, McAleer, Fried, and Salakhutdinov}]{koh2024tree}
Jing~Yu Koh, Stephen McAleer, Daniel Fried, and Ruslan Salakhutdinov. 2024{\natexlab{b}}.
\newblock Tree search for language model agents.
\newblock \emph{arXiv preprint arXiv:2407.01476}.

\bibitem[{Krishnamurthy et~al.(2024)Krishnamurthy, Harris, Foster, Zhang, and Slivkins}]{krishnamurthy2024can}
Akshay Krishnamurthy, Keegan Harris, Dylan~J Foster, Cyril Zhang, and Aleksandrs Slivkins. 2024.
\newblock Can large language models explore in-context?
\newblock \emph{arXiv preprint arXiv:2403.15371}.

\bibitem[{Laskin et~al.(2022)Laskin, Wang, Oh, Parisotto, Spencer, Steigerwald, Strouse, Hansen, Filos, Brooks et~al.}]{laskin2022context}
Michael Laskin, Luyu Wang, Junhyuk Oh, Emilio Parisotto, Stephen Spencer, Richie Steigerwald, DJ~Strouse, Steven Hansen, Angelos Filos, Ethan Brooks, et~al. 2022.
\newblock In-context reinforcement learning with algorithm distillation.
\newblock \emph{arXiv preprint arXiv:2210.14215}.

\bibitem[{Lee et~al.(2024)Lee, Xie, Pacchiano, Chandak, Finn, Nachum, and Brunskill}]{lee2024supervised}
Jonathan Lee, Annie Xie, Aldo Pacchiano, Yash Chandak, Chelsea Finn, Ofir Nachum, and Emma Brunskill. 2024.
\newblock Supervised pretraining can learn in-context reinforcement learning.
\newblock \emph{Advances in Neural Information Processing Systems}, 36.

\bibitem[{Li et~al.(2024)Li, Zhang, Zhang, Guo, Zhang, Li, Zhang, Liu, and Li}]{li2024llavanext-strong}
Bo~Li, Kaichen Zhang, Hao Zhang, Dong Guo, Renrui Zhang, Feng Li, Yuanhan Zhang, Ziwei Liu, and Chunyuan Li. 2024.
\newblock \href {https://llava-vl.github.io/blog/2024-05-10-llava-next-stronger-llms/} {Llava-next: Stronger llms supercharge multimodal capabilities in the wild}.

\bibitem[{Lin et~al.(2024)Lin, Li, Gao, Yang, Wu, Bai, Lei, Wang, and Shou}]{lin2024showui}
Kevin~Qinghong Lin, Linjie Li, Difei Gao, Zhengyuan Yang, Shiwei Wu, Zechen Bai, Weixian Lei, Lijuan Wang, and Mike~Zheng Shou. 2024.
\newblock Showui: One vision-language-action model for gui visual agent.
\newblock \emph{arXiv preprint arXiv:2411.17465}.

\bibitem[{Liu et~al.(2024)Liu, Ou, Song, Qu, Lam, Xiong, Chen, Neubig, and Yue}]{liu2024harnessing}
Junpeng Liu, Tianyue Ou, Yifan Song, Yuxiao Qu, Wai Lam, Chenyan Xiong, Wenhu Chen, Graham Neubig, and Xiang Yue. 2024.
\newblock Harnessing webpage uis for text-rich visual understanding.
\newblock \emph{arXiv preprint arXiv:2410.13824}.

\bibitem[{Lu et~al.(2024)Lu, Yang, Shen, and Awadallah}]{lu2024omniparserpurevisionbased}
Yadong Lu, Jianwei Yang, Yelong Shen, and Ahmed Awadallah. 2024.
\newblock \href {https://arxiv.org/abs/2408.00203} {Omniparser for pure vision based gui agent}.
\newblock \emph{Preprint}, arXiv:2408.00203.

\bibitem[{Monea et~al.(2024)Monea, Bosselut, Brantley, and Artzi}]{monea2024llms}
Giovanni Monea, Antoine Bosselut, Kiant{\'e} Brantley, and Yoav Artzi. 2024.
\newblock Llms are in-context reinforcement learners.
\newblock \emph{arXiv preprint arXiv:2410.05362}.

\bibitem[{OpenAI(2024)}]{gpt4o}
OpenAI. 2024.
\newblock \href {https://openai.com/index/hello-gpt-4o/} {Gpt-4o}.

\bibitem[{Pan(2023)}]{pan2023context}
Jane Pan. 2023.
\newblock What in-context learning “learns” in-context: Disentangling task recognition and task learning.
\newblock Master's thesis, Princeton University.

\bibitem[{Sodhi et~al.(2024)Sodhi, Branavan, Artzi, and McDonald}]{sodhi2024step}
Paloma Sodhi, SRK Branavan, Yoav Artzi, and Ryan McDonald. 2024.
\newblock Step: Stacked llm policies for web actions.
\newblock In \emph{First Conference on Language Modeling}.

\bibitem[{Wang et~al.(2023)Wang, Li, Dai, Chen, Zhou, Meng, Zhou, and Sun}]{wang2023label}
Lean Wang, Lei Li, Damai Dai, Deli Chen, Hao Zhou, Fandong Meng, Jie Zhou, and Xu~Sun. 2023.
\newblock Label words are anchors: An information flow perspective for understanding in-context learning.
\newblock \emph{arXiv preprint arXiv:2305.14160}.

\bibitem[{Wei et~al.(2023)Wei, Wei, Tay, Tran, Webson, Lu, Chen, Liu, Huang, Zhou et~al.}]{wei2023larger}
Jerry Wei, Jason Wei, Yi~Tay, Dustin Tran, Albert Webson, Yifeng Lu, Xinyun Chen, Hanxiao Liu, Da~Huang, Denny Zhou, et~al. 2023.
\newblock Larger language models do in-context learning differently.
\newblock \emph{arXiv preprint arXiv:2303.03846}.

\bibitem[{Wu et~al.(2024)Wu, Wu, Xu, Wang, Sun, Jia, Cheng, Ding, Chen, Liang et~al.}]{wu2024atlas}
Zhiyong Wu, Zhenyu Wu, Fangzhi Xu, Yian Wang, Qiushi Sun, Chengyou Jia, Kanzhi Cheng, Zichen Ding, Liheng Chen, Paul~Pu Liang, et~al. 2024.
\newblock Os-atlas: A foundation action model for generalist gui agents.
\newblock \emph{arXiv preprint arXiv:2410.23218}.

\bibitem[{Xie et~al.(2024)Xie, Zhang, Chen, Li, Zhao, Cao, Hua, Cheng, Shin, Lei, Liu, Xu, Zhou, Savarese, Xiong, Zhong, and Yu}]{OSWorld}
Tianbao Xie, Danyang Zhang, Jixuan Chen, Xiaochuan Li, Siheng Zhao, Ruisheng Cao, Toh~Jing Hua, Zhoujun Cheng, Dongchan Shin, Fangyu Lei, Yitao Liu, Yiheng Xu, Shuyan Zhou, Silvio Savarese, Caiming Xiong, Victor Zhong, and Tao Yu. 2024.
\newblock \href {https://arxiv.org/abs/2404.07972} {Osworld: Benchmarking multimodal agents for open-ended tasks in real computer environments}.
\newblock \emph{Preprint}, arXiv:2404.07972.

\bibitem[{Xu et~al.(2022)Xu, Shen, Zhang, Lu, Zhao, Tenenbaum, and Gan}]{xu2022prompting}
Mengdi Xu, Yikang Shen, Shun Zhang, Yuchen Lu, Ding Zhao, Joshua Tenenbaum, and Chuang Gan. 2022.
\newblock Prompting decision transformer for few-shot policy generalization.
\newblock In \emph{international conference on machine learning}, pages 24631--24645. PMLR.

\bibitem[{Yang et~al.(2024)Yang, Yang, Hui, Zheng, Yu, Zhou, Li, Li, Liu, Huang, Dong, Wei, Lin, Tang, Wang, Yang, Tu, Zhang, Ma, Xu, Zhou, Bai, He, Lin, Dang, Lu, Chen, Yang, Li, Xue, Ni, Zhang, Wang, Peng, Men, Gao, Lin, Wang, Bai, Tan, Zhu, Li, Liu, Ge, Deng, Zhou, Ren, Zhang, Wei, Ren, Fan, Yao, Zhang, Wan, Chu, Liu, Cui, Zhang, and Fan}]{qwen2}
An~Yang, Baosong Yang, Binyuan Hui, Bo~Zheng, Bowen Yu, Chang Zhou, Chengpeng Li, Chengyuan Li, Dayiheng Liu, Fei Huang, Guanting Dong, Haoran Wei, Huan Lin, Jialong Tang, Jialin Wang, Jian Yang, Jianhong Tu, Jianwei Zhang, Jianxin Ma, Jin Xu, Jingren Zhou, Jinze Bai, Jinzheng He, Junyang Lin, Kai Dang, Keming Lu, Keqin Chen, Kexin Yang, Mei Li, Mingfeng Xue, Na~Ni, Pei Zhang, Peng Wang, Ru~Peng, Rui Men, Ruize Gao, Runji Lin, Shijie Wang, Shuai Bai, Sinan Tan, Tianhang Zhu, Tianhao Li, Tianyu Liu, Wenbin Ge, Xiaodong Deng, Xiaohuan Zhou, Xingzhang Ren, Xinyu Zhang, Xipin Wei, Xuancheng Ren, Yang Fan, Yang Yao, Yichang Zhang, Yu~Wan, Yunfei Chu, Yuqiong Liu, Zeyu Cui, Zhenru Zhang, and Zhihao Fan. 2024.
\newblock Qwen2 technical report.
\newblock \emph{arXiv preprint arXiv:2407.10671}.

\bibitem[{Yang et~al.(2023)Yang, Zhang, Li, Zou, Li, and Gao}]{yang2023setofmark}
Jianwei Yang, Hao Zhang, Feng Li, Xueyan Zou, Chunyuan Li, and Jianfeng Gao. 2023.
\newblock Set-of-mark prompting unleashes extraordinary visual grounding in gpt-4v.
\newblock \emph{arXiv preprint arXiv:2310.11441}.

\bibitem[{You et~al.(2024)You, Zhang, Schoop, Weers, Swearngin, Nichols, Yang, and Gan}]{you2024ferret}
Keen You, Haotian Zhang, Eldon Schoop, Floris Weers, Amanda Swearngin, Jeffrey Nichols, Yinfei Yang, and Zhe Gan. 2024.
\newblock Ferret-ui: Grounded mobile ui understanding with multimodal llms.
\newblock \emph{arXiv preprint arXiv:2404.05719}.

\bibitem[{Zheng et~al.(2024{\natexlab{a}})Zheng, Gou, Kil, Sun, and Su}]{zheng2024gpt}
Boyuan Zheng, Boyu Gou, Jihyung Kil, Huan Sun, and Yu~Su. 2024{\natexlab{a}}.
\newblock Gpt-4v (ision) is a generalist web agent, if grounded.
\newblock \emph{arXiv preprint arXiv:2401.01614}.

\bibitem[{Zheng et~al.(2024{\natexlab{b}})Zheng, Gou, Kil, Sun, and Su}]{zheng2023seeact}
Boyuan Zheng, Boyu Gou, Jihyung Kil, Huan Sun, and Yu~Su. 2024{\natexlab{b}}.
\newblock Gpt-4v(ision) is a generalist web agent, if grounded.
\newblock \emph{arXiv preprint arXiv:2401.01614}.

\bibitem[{Zhou et~al.(2023)Zhou, Xu, Zhu, Zhou, Lo, Sridhar, Cheng, Ou, Bisk, Fried et~al.}]{zhou2023webarena}
Shuyan Zhou, Frank~F Xu, Hao Zhu, Xuhui Zhou, Robert Lo, Abishek Sridhar, Xianyi Cheng, Tianyue Ou, Yonatan Bisk, Daniel Fried, et~al. 2023.
\newblock Webarena: A realistic web environment for building autonomous agents.
\newblock \emph{arXiv preprint arXiv:2307.13854}.

\end{thebibliography}

\appendix

\label{sec:appendix}
\newpage

\section{GUI Elements Fuzzy Visual Matching}
\label{Appendix_fuzz_matching}
We develop a GUI Elements Fuzzy Visual Matching module $F_{fvm}$, to compare if two GUI elements can be recognized as visually the same. 
Challenges arise from the variations in GUI rendering; for example, web browsers could render the sample page with slight element shifts each time. Such variation can make pixel-perfect matching overly sensitive, leading to false negatives. Furthermore, dynamic elements on the screen, such as GIFs, can also cause variability unrelated to the executed action.
Our solution is let $F_{fvm}$ compares $e$ to the corresponding patch $e'$ cropped at the same location in \(i^{t+1}\) for each GUI element $e$ in \(i^t\), and output a difference score $p = Max(F_{fvm}(e, e')), e \in i^t, e' \in T^{t+1}$. Specifically, a Gaussian filter is first applied to each pair of $e$ and $e'$ to smooth rendering defects. Then $e$ and $e'$ are aligned with varying shifts, ensuring a minimum overlap of 75\%, to compute the maximum normalized pixel-wise difference $F_{fvm}(e, e')$. Finally, based on the $F_{fvm}(e, e')$, $e$ and $e'$ are considered identical if $F_{fvm}(e, e') \leq 0.05$. 

The GUI Elements Fuzzy Visual Matching module is not only used for retrieving in-context examples $a_{eg1}$ and $a_{eg2}$ as mentioned in Section \ref{incontextAS}, it is also crucial for verifying whether the executed action $a^t$ transitions \(i^t\) to a new screen \(i^{t+1}\), and whether \(i^{t+1}\) is already in the exploration graph, so that the nodes in the graph are unique.
\(i^{t+1}\) is regarded the same as \(i^t\) if $p \leq 0.05$. Additionally, for dynamic content, $F_{fvm}$ compares the same elements across multiple screenshots over time from the same screen to identify inconsistent regions and excludes them when calculating $p$.

\section{Generating GUI Action Grounding Queries ($u^t$)}  
\label{Appendix_prompt_for_u}
\begin{figure*}[t] 
\includegraphics[width=\textwidth]{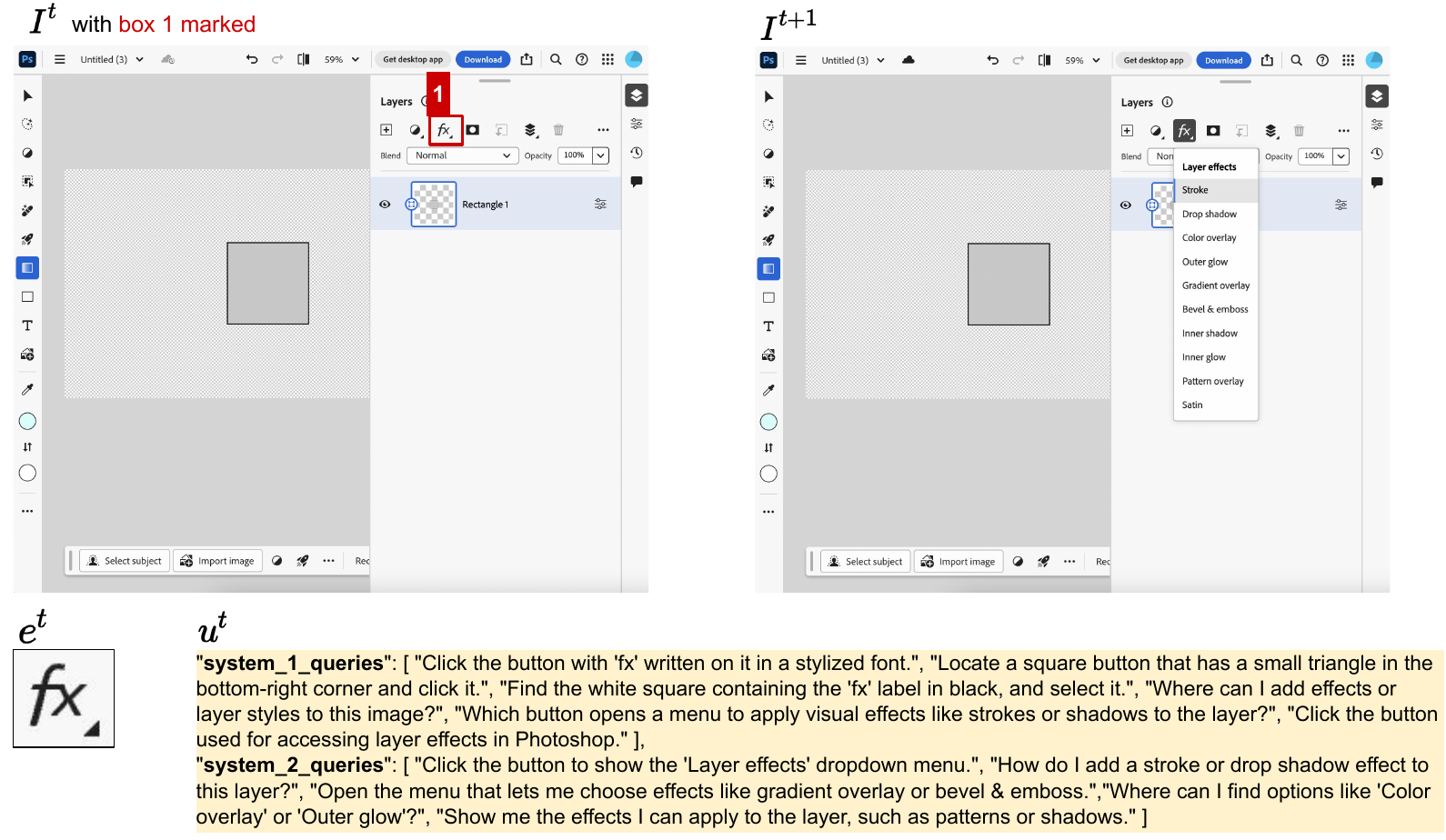}
\caption{Examples of the input images ($i^t$ with box 1 marked, $i^{t+1}$ and $e^t$) and output GUI action grounding queries $u^t$ in process of generating data generation along the exploration.}
\label{prompt1_output}
\end{figure*}

Once a new edge \((i^t, a^t, i^{t+1})\) is added during exploration, we send this information to the MLLM to generate $u^t$, a list of action grounding queries for the target element \(e^t\) in \(i^t\). The generation process uses a carefully crafted prompt, designed to ensure the queries cover both System 1 (focused on current screen content) and System 2 (anticipating interaction outcomes) grounding challenges.  
The full version of the text prompt for the MLLM is provided in Figure \ref{prompt1}. We also show an example of the input images in Figure \ref{prompt1_output} along with the GUI action grounding queries $u^t$ in the correspnding output.

\begin{figure}[h] 
\begin{smallermdframed} 
A use clicked the element marked with box 1 on the screen shown in the first image, and then arrived at the screen shown in the second image. A zoomed in look of the element clicked in shown in the third image. Please generate a json dictionary format with values for the following 3 keys:

1. analysis: describing the appearance of the element, including but not limited to color, shape, etc. Try to make the description uniquely identify the element.

2. system\_1\_queries: a list of maximum 6 requests or questions that will uniquely lead to clicking the element in the page, with maximum 3 of them mentioning something special about the appearance of the element (can be skipped if the element's appearance is just plain text).

3. system\_2\_queries: a list of maximum 5 simple requests or questions that uniquely lead to the element. They each should mention one specific function (consequence) of this click, i.e. a specific thing that is only shown in the second image, but not in the first image.
\end{smallermdframed}
\caption{Text prompt for generating GUI action grounding queries ($u^t$).}
\label{prompt1}
\end{figure}


  
    
    
  
  
    
    
    
    
    
    
  
  
    
    
    
    
    
  


\section{Approximating State ($S^t$) with Natural Language Descriptions $D^t$}
\label{Appendix_approx_s_with_d}
To simplify the representation of the state \(S^t\) at the \(t\)-th exploration step, we approximate it with a list of natural language descriptions \(D^t\), where each description \(d^k\) corresponds to an action and its resulting state transition. Figure \ref{prompt:prompt3} illustrates the input prompt used to generate one such natural language description. The input consists of the current screen \(i^t\) with the action target \(a^t\) visually marked (box 1), along with the resulting screen \(i^{t+1}\). Using this input, the MLLM produces a textual description capturing the key details of the transition, including the action \(a^t\), the visual changes between \(i^t\) and \(i^{t+1}\), and any notable observations.
These natural language descriptions serve as a compact and interpretable representation of the exploration history, enabling efficient input to the MLLM during subsequent steps of the Q-ICRL process.

\begin{figure}[h] 
\begin{smallermdframed} 
A use clicked an element on the screen shown in the first image, and then arrived at the screen shown in the second image.
Your output should be a json dictionary format with values for the following 2 keys:

1. consequence: what happens after the click and what is shown based on the second image.

2. clicked\_element: describe what element (marked by the box 1) is clicked (apperance, layout, etc).

Note: box 1 is the bounding box with label 1.
Note: do not mention box 1 in your output.

\end{smallermdframed}
\caption{Text prompt for generating a natural language description $d^t$ of one exploration step at $t$, where the input images are $i^t$ with box 1 marked and $i^{t+1}$.}
\label{prompt:prompt3}
\end{figure}

\section{Predicting \(\hat{Q}(a^t_x)\) with MLLM through In-Context Learning}  
\label{Appendix_ICL}
\begin{figure*}[t]
    \centering
    \includegraphics[width=\textwidth]{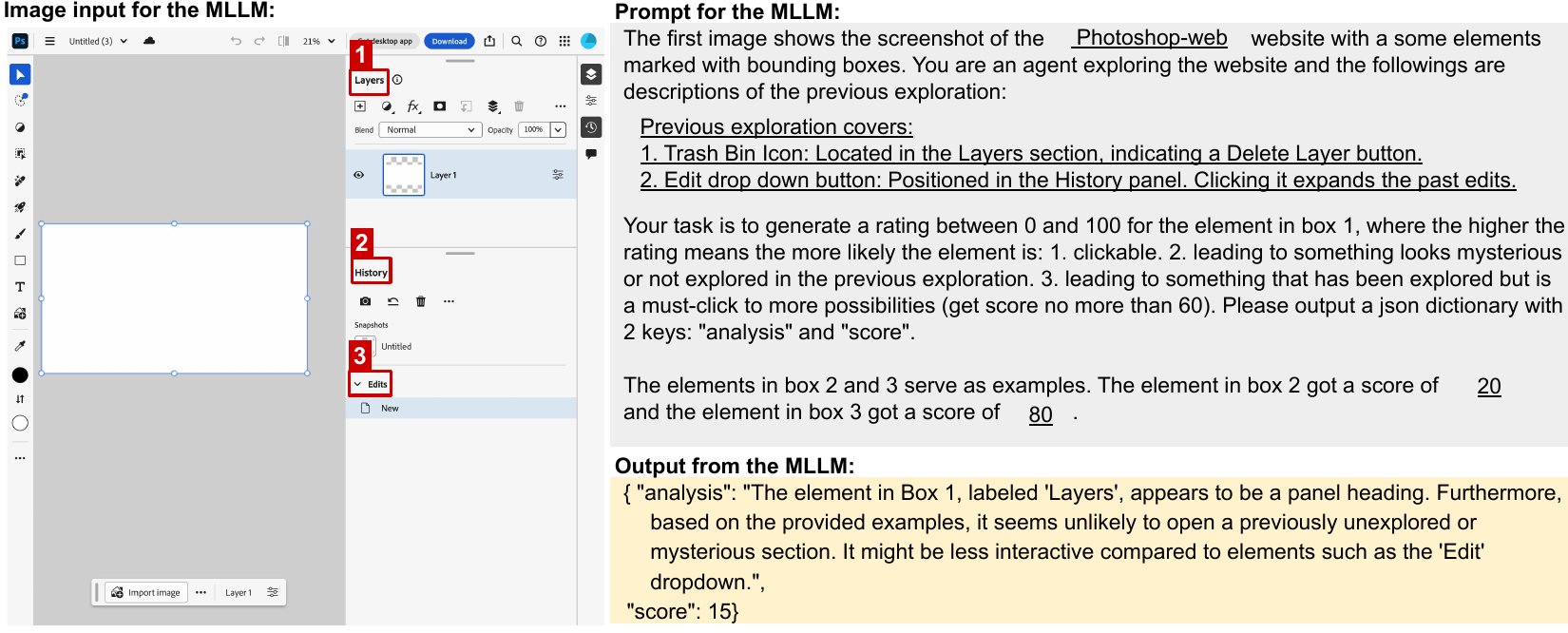}
    \caption{Example of the full MLLM input and output when predicting \(\hat{Q}(a^t_x)\) through in-context learning. The input includes two example actions \((a_{\text{eg1}}, a_{\text{eg2}})\) marked by bounding boxes 2 and 3, and the candidate action \(a^t_x\) marked by bounding box 1. The prompt if formed by a fixed template with the GUI environment name, state \(S^t\), and \((Q(a_{\text{eg1}}), Q(a_{\text{eg2}}))\) that are all underlined. The output is the predicted Q-value \(\hat{Q}(a^t_x)\) for \(a^t_x\).}    
    \label{fig_appendix_icl}
\end{figure*}

To predict \(\hat{Q}(a^t_x)\) for candidate actions \(a^t_x\), in the text prompt, we include the MLLM with the natural language escriptions $D^t$ approximating current state \(S^t\), two example actions \((a_{\text{eg1}}, a_{\text{eg2}})\) and their Q-values \((Q(a_{\text{eg1}}), Q(a_{\text{eg2}}))\), along with the visual input using the Set-of-Mark method \cite{yang2023setofmark}. Figure \ref{fig_appendix_icl} shows an example of the full input and output.

\section{Experiment Details for Exploration and Fine-tuning}  
\label{appendix_exp_d}
\paragraph{Exploration Configurations}  
During exploration, the GUI-Bee agent operates with a maximum of \(T = 400\) exploration steps, and at each step, it samples \(H = 3\) candidate actions. The action type \(c_t\) is restricted to "click" or "scroll" transitions, as these are the most common actions for GUI navigation. For "scroll" actions, the target element \(e^t\) is simplified to represent the "full page," ensuring consistent representation of scroll transitions. To enhance robustness, each environment is explored three times using different screen resolutions. This variation ensures the generated data captures diverse screen setups, improving the generalization ability of the fine-tuned models. Each exploration session lasts between 6 to 18 hours, depending on web loading latency. The long loading time is due to the overhead of using the Playwright tool \footnote{https://playwright.dev} to acquire the accessibility tree for each screen, which can be further optimized with some engineering efforts. 

\paragraph{Model Fine-tuning Configurations}  
We fine-tune three GUI grounding models—SeeClick \cite{cheng2024seeclick}, Qwen-GUI \cite{chen2024guicource}, and UIX-7B \cite{liu2024harnessing}—using the data generated by the GUI-Bee agent. Fine-tuning is performed in two input configurations: vision-only, where the input consists of GUI screenshots only, and Vision+A11y, where the input includes both GUI screenshots and the accessibility tree embedded in the text prompt.  

The accessibility tree (A11y tree) is a structured representation of the GUI that exposes key information about screen elements, such as their type, properties, and hierarchical relationships. Typically used for assistive technologies like screen readers, the A11y tree provides textual descriptions and spatial information of the interface components, complementing visual input for models. Including this information in the input prompt allows models to leverage both visual and structural cues, improving their grounding accuracy.  

SeeClick and Qwen-GUI are based on Qwen-VL \cite{bai2023qwen}, while UIX-7B is derived from Llava-1.6 \cite{li2024llavanext-strong} with Qwen2-7B-Instruct \cite{qwen2} as the primary LLM backbone. For models that predict bounding boxes, such as Qwen-GUI and UIX-7B, the center of the predicted bounding box is used as the final output point for evaluation.

\paragraph{Fine-tuning Settings}  
For all models, fine-tuning is conducted with a batch size of 16, a learning rate of \(1 \times 10^{-6}\), and for 3 training epochs. The generated exploration data are formatted to match the original training format of these models to ensure consistency. For models trained with bounding boxes, the ground truth bounding box coordinates are converted to center points to align with evaluation requirements.

\paragraph{Computation Time}  
Each model is fine-tuned for 3 epochs. The overall fine-tuning time depends on the model and the input configuration. Configurations using both screenshots and accessibility tree information require slightly longer processing time due to the additional textual input.

\section{Data Annotation Details}  
\label{appendix-annotation}
We recruited annotators from within our research team, ensuring familiarity with GUI environments. Annotators were compensated fairly for their work to maintain ethical standards. All queries undergo manual review to eliminate ambiguity, duplication, and low quality, ensuring alignment with recorded actions.

\begin{figure*}[t]
    \centering
    \includegraphics[width=\linewidth]{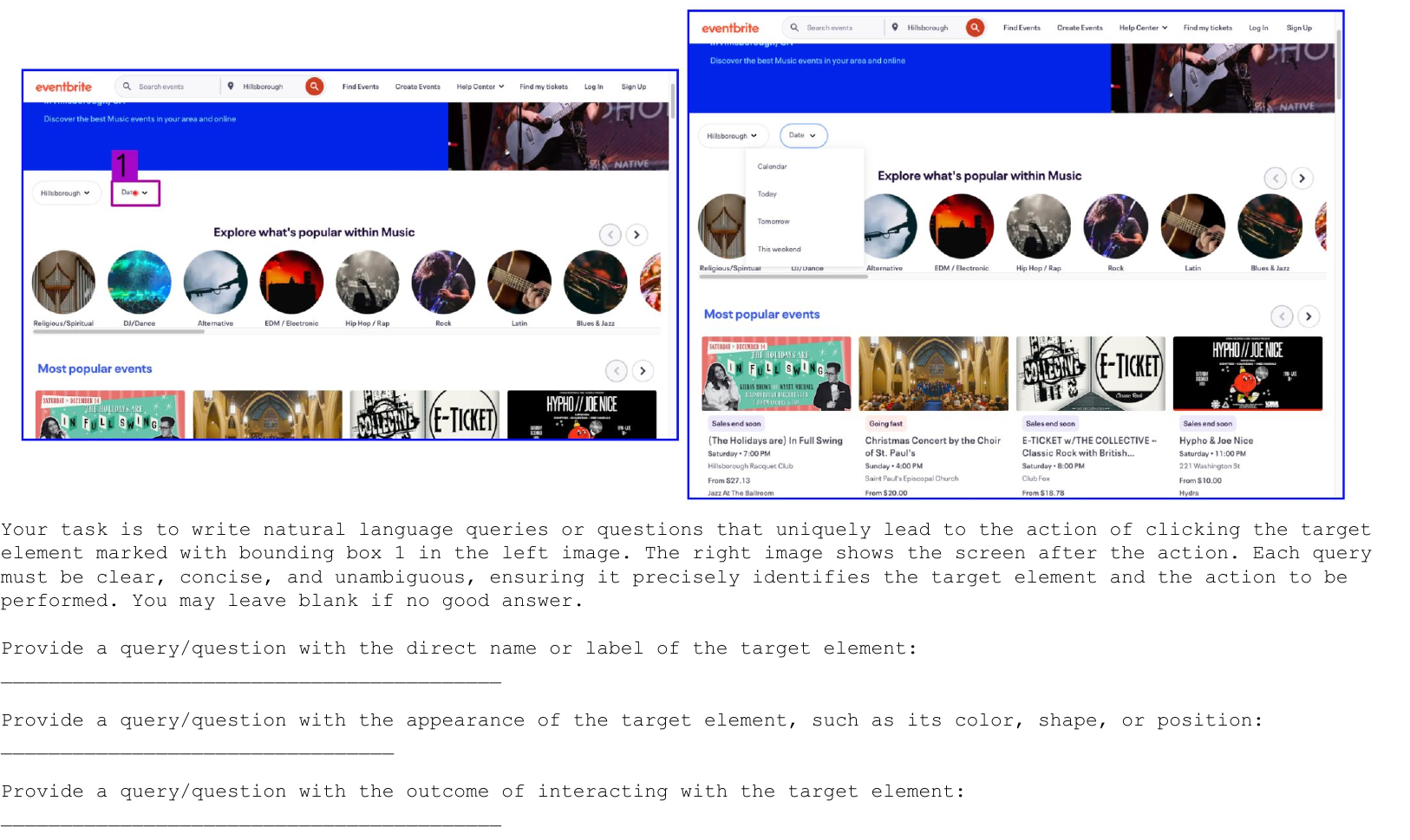}
    \caption{Example of the annotation interface for collecting GUI action grounding queries. The target element is marked with bounding box 1, and annotators will write queries uniquely identifying this action.}
    \label{fig:anno_interface}
\end{figure*}

\paragraph{Annotation Interface}  
Figure \ref{fig:anno_interface} shows an example of the annotation interface, where annotators view the screens and marked target element to input queries efficiently.  

\section{Examples of NovelScreenSpot}  
\label{appendix_NSS_example}
\begin{figure*}[t]
    \centering
    \includegraphics[width=0.9\linewidth]{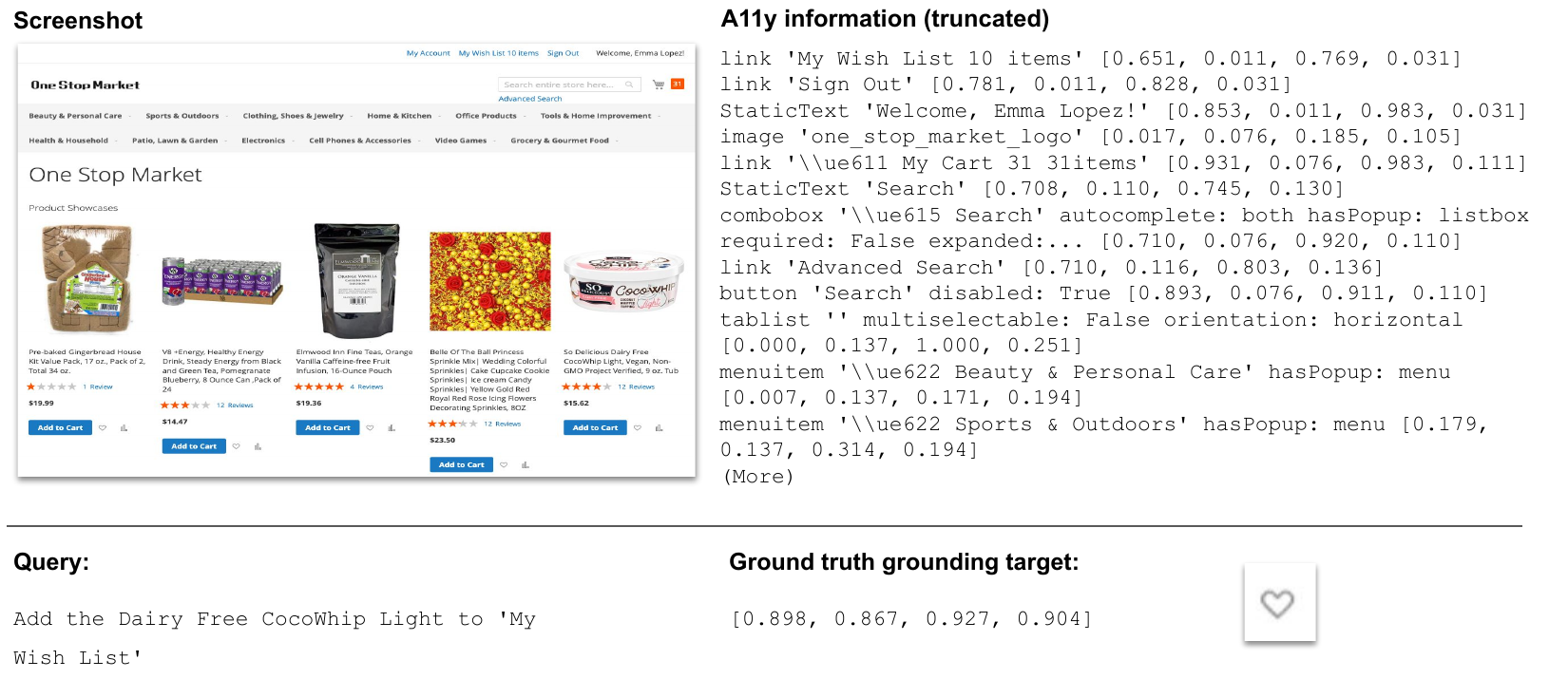}
    \caption{Example of NovelScreenSpot data from the Shopping environment.}
    \label{fig:shopping}
\end{figure*}

\begin{figure*}[t]
    \centering
    \includegraphics[width=0.9\linewidth]{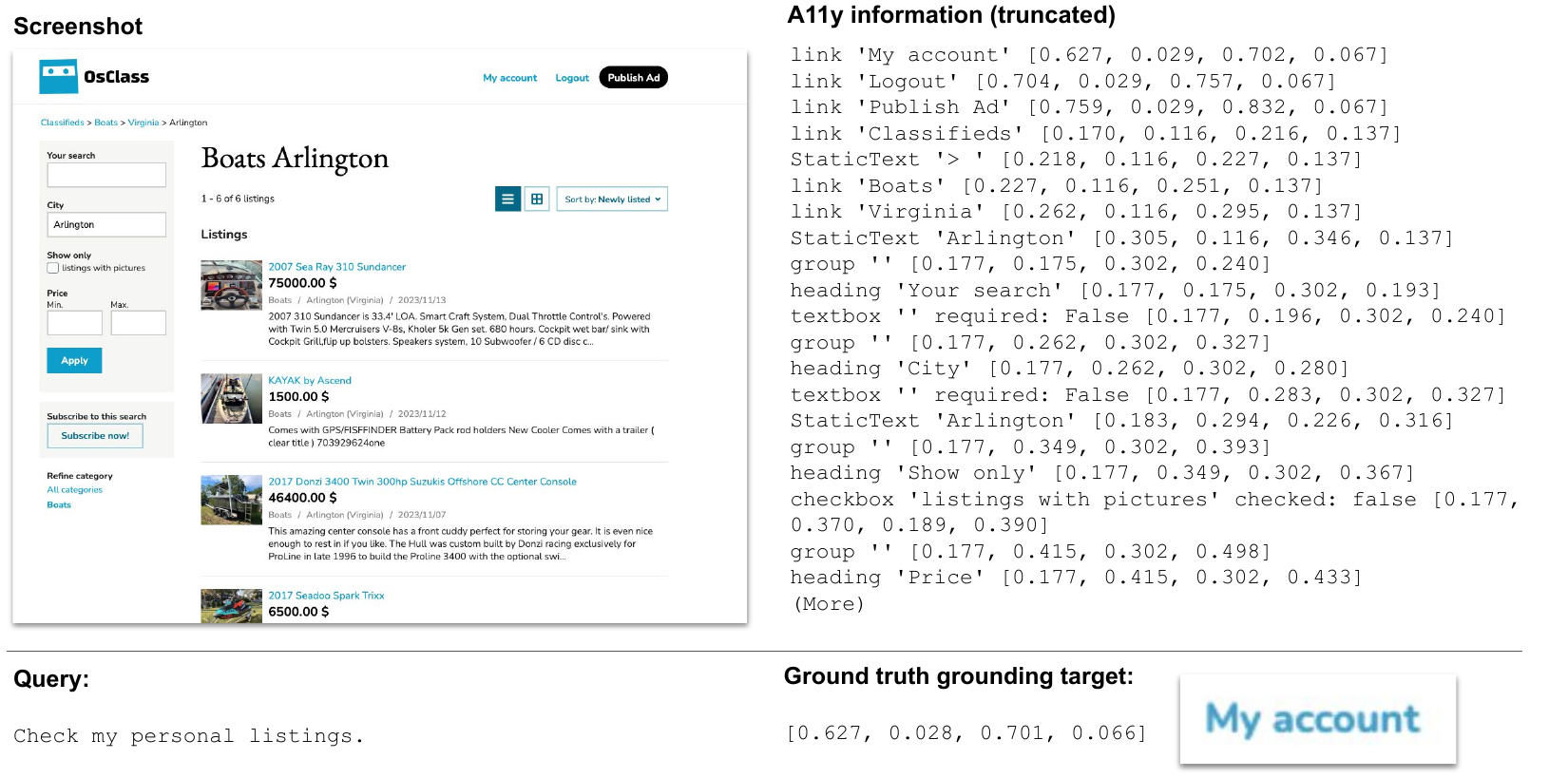}
    \caption{Example of NovelScreenSpot data from the Classifieds environment.}
    \label{fig:classi}
\end{figure*}

\begin{figure*}[t]
    \centering
    \includegraphics[width=0.9\linewidth]{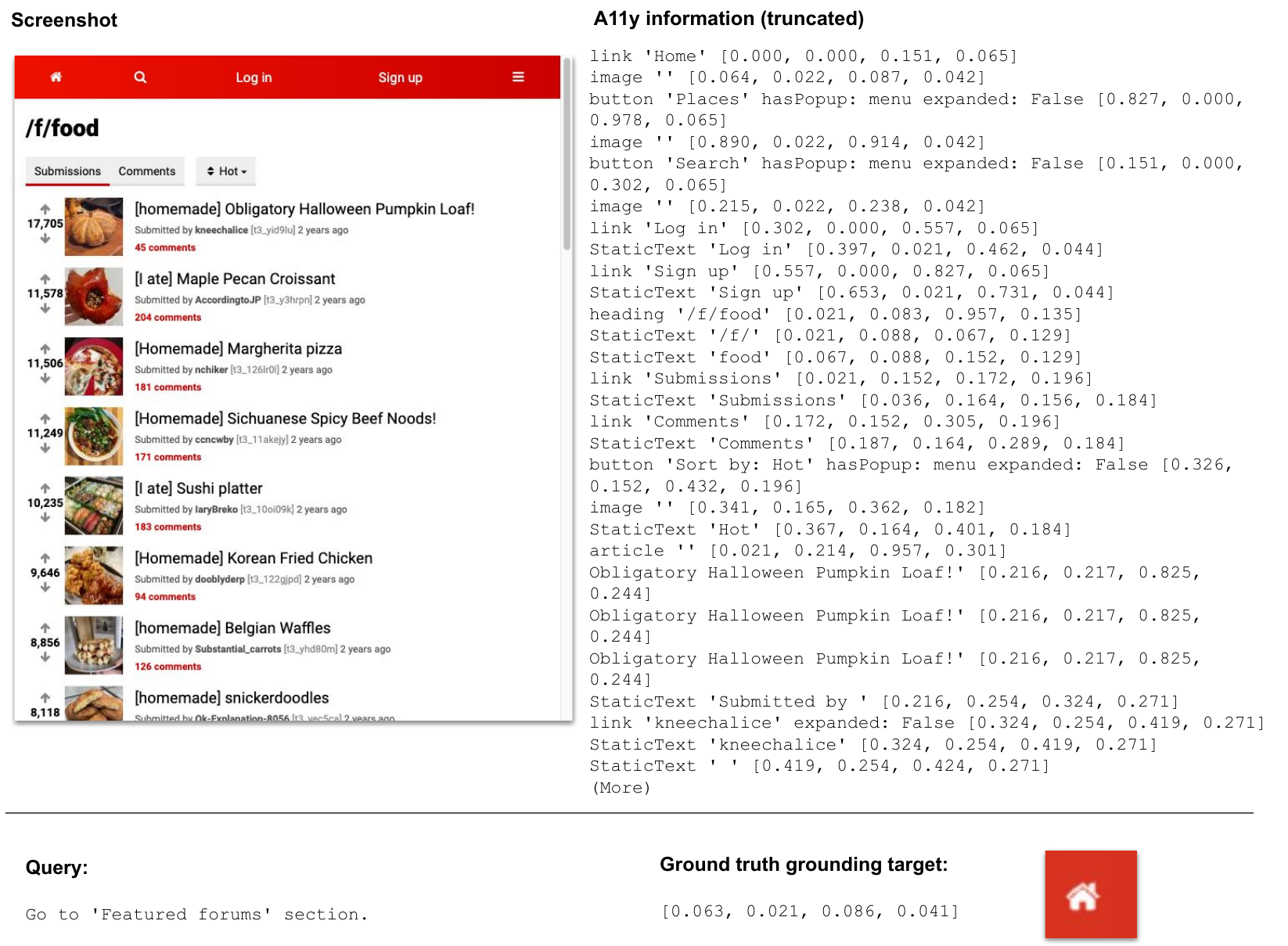}
    \caption{Example of NovelScreenSpot data from the Reddit environment.}
    \label{fig:reddit}
\end{figure*}

\begin{figure*}[t]
    \centering
    \includegraphics[width=0.9\linewidth]{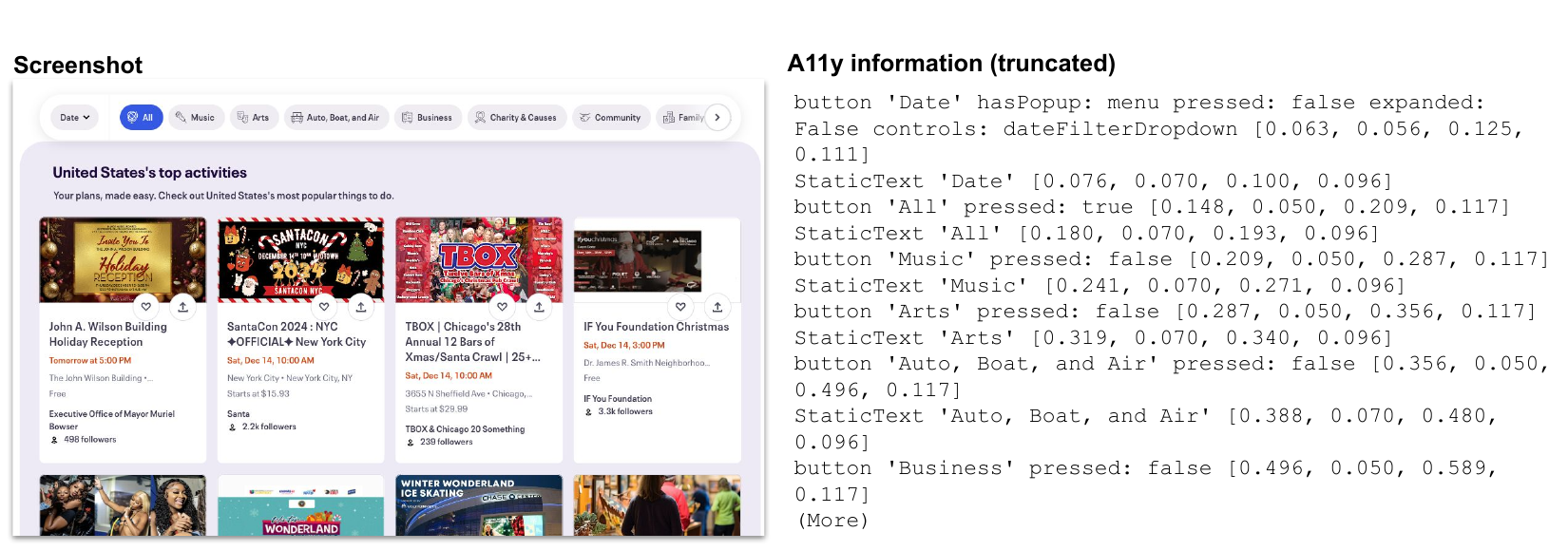}
    \caption{Example of NovelScreenSpot data from the Eventbrite environment.}
    \label{fig:event}
\end{figure*}

\begin{figure*}[t]
    \centering
    \includegraphics[width=0.9\linewidth]{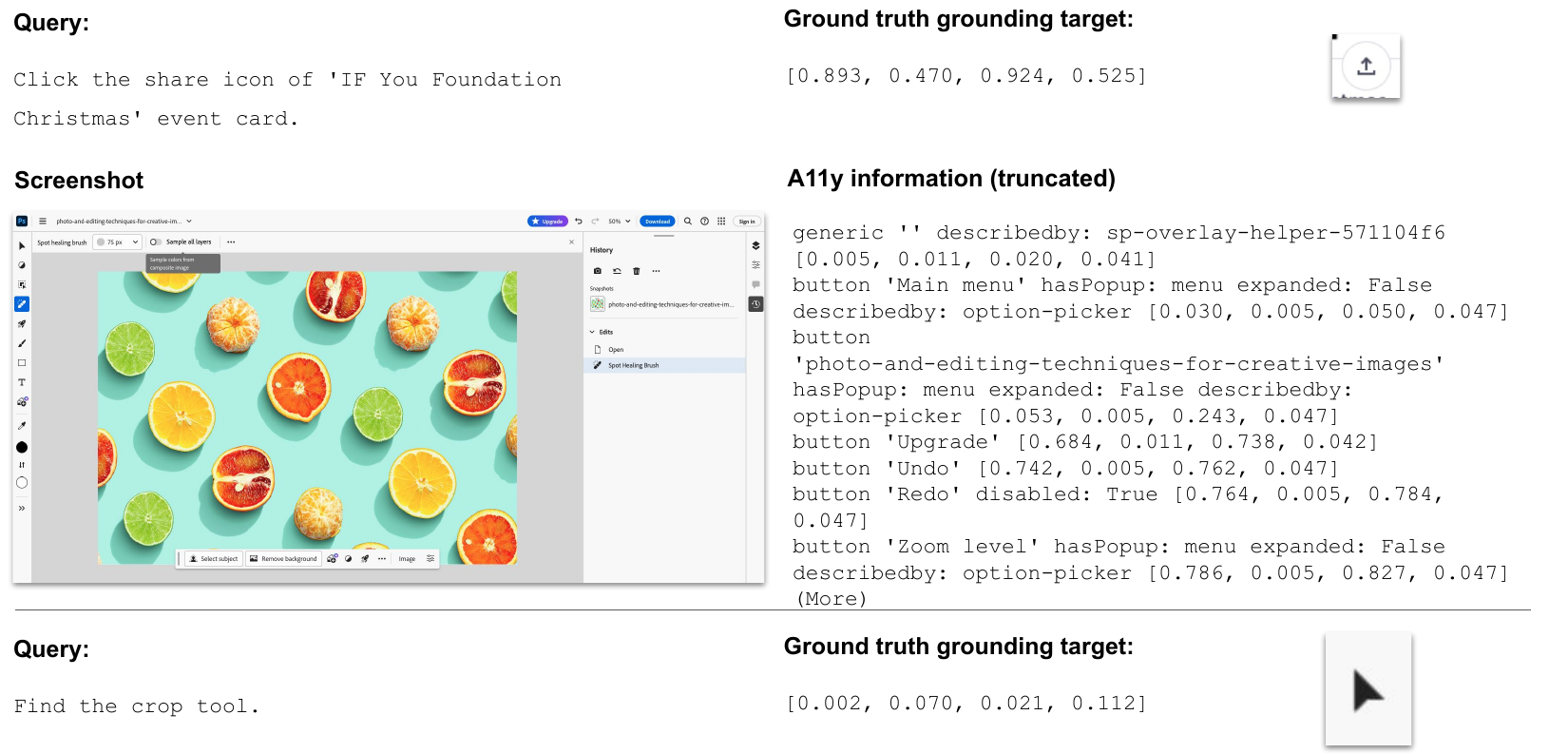}
    \caption{Example of NovelScreenSpot data from the Photoshop-web environment.}
    \label{fig:photo}
\end{figure*}

We randomly sample data from each environment in the NovelScreenSpot benchmark and present examples in Figure \ref{fig:shopping}, \ref{fig:classi}, \ref{fig:reddit}, \ref{fig:event}, and \ref{fig:photo}. Each figure illustrates the GUI screen, the A11y string, the corresponding query, and the ground truth target element, showcasing the diversity and environment-specific nature of the benchmark.

\end{document}